\documentclass[preprint,12pt]{elsarticle}
\usepackage[x11names]{xcolor}
\usepackage{amsmath}	
\usepackage[colorlinks]{hyperref}
\usepackage[capitalise]{cleveref}
\usepackage{siunitx} 
\usepackage{float}

\usepackage[nolist]{acronym}

\usepackage{tabularx}

\definecolor{kitgruen}{RGB}{0 ,150,130}
\definecolor{kitblau}{RGB}{70,100,170}
\definecolor{kitschwarz}{RGB}{0,0,0}
\definecolor{kitgrau}{RGB}{64,64,64}

\definecolor{kitgelb}{RGB}{ 252,229,0}
\definecolor{kitorange}{RGB}{223,155,27}
\definecolor{kitmaigruen}{RGB}{ 140,182,60}
\definecolor{kitrot}{RGB}{162 ,34,35}
\definecolor{kitlila}{RGB}{163,16,124}
\definecolor{kitbraun}{RGB}{167,130,49}
\definecolor{kitcyan}{RGB}{35,161,224}

\usepackage{amssymb}
\usepackage{cite}

\journal{CMAME}

\begin{document}
\begin{acronym}[mylongestacronym]
\acro{gns}[GNS]{graph network simulator}
\acro{cnn}[CNN]{convolutional neural network}
\acro{gnn}[GNN]{graph neural network}
\acro{gccn}[GCCN]{graph Chebyshev convolutional neural network}
\acro{piggn}[PI-GGN]{physics-informed graph neural Galerkin network}
\acro{fem}[FEM]{finite element method}
\acro{fdm}[FDM]{finite difference method }
\acro{fvm}[FVM]{finite volume method}
\acro{bem}[BEM]{boundary element method}
\acro{pde}[PDE]{partial differential equation}
\acro{cad}[CAD]{computer aided design}
\acro{pinn}[PINN]{physics-informed neural network}
\acro{iga}[IGA]{isogeometric analysis}
\acro{pigns}[PI-GNS]{physics-informed graph network simulator}
\acro{ddgns}[DD-GNS]{data-driven graph network simulator}
\acro{mgn}[MGN]{\textsc{MeshGraphNet}}
\acrodefplural{mgn}[MGNs]{\textsc{MeshGraphNets}}
\acro{ddmgn}[DD-MGN]{data-driven \textsc{MeshGraphNet}}
\acrodefplural{ddmgn}[DD-MGNs]{data-driven \textsc{MeshGraphNets}}
\acro{pimgn}[PI-MGN]{physics-informed \textsc{MeshGraphNet}}
\acrodefplural{pimgn}[PI-MGNs]{physics-informed \textsc{MeshGraphNets}}
\acro{ml}[ML]{machine learning}
\acro{mpn}[MPN]{message passing network}
\acro{mlp}[MLP]{multilayer perceptron}
\acro{mse}[MSE]{mean squared error}
\acro{relu}[ReLU]{rectified linear unit}
\end{acronym}

\begin{frontmatter}



\title{Physics-informed \textsc{MeshGraphNets} (PI-MGNs): Neural finite element solvers for non-stationary and nonlinear simulations on arbitrary meshes}


\author[affiliation1]{Tobias Würth}
\author[affiliation2]{Niklas Freymuth}
\author[affiliation1]{Clemens Zimmerling}
\author[affiliation2]{Gerhard Neumann}
\author[affiliation1]{Luise Kärger}

\affiliation[affiliation1]{organization={Karlsruhe Institute of Technology (KIT), Institute of Vehicle System Technology},
            city={Karlsruhe},
            country={Germany}}
\affiliation[affiliation2]{organization={Karlsruhe Institute of Technology (KIT), Autonomous Learning Robots},
            city={Karlsruhe},
            country={Germany}}

\begin{abstract}
Engineering components must meet increasing technological demands in ever shorter development cycles. To face these challenges, a holistic approach is essential that allows for the concurrent development of part design, material system and manufacturing process. Current approaches employ numerical simulations, which however quickly becomes computation-intensive, especially for iterative optimization. Data-driven machine learning methods can be used to replace time- and resource-intensive numerical simulations. In particular, \acp{mgn} have shown promising results. They enable fast and accurate predictions on unseen mesh geometries while being fully differentiable for optimization. However, these models rely on large amounts of expensive training data, such as numerical simulations. \Acp{pinn} offer an opportunity to train neural networks with partial differential equations instead of labeled data, but have not been extended yet to handle time-dependent simulations of arbitrary meshes. This work introduces \acsp{pimgn}, a hybrid approach that combines \acp{pinn} and \acp{mgn} to quickly and accurately solve non-stationary and nonlinear \acp{pde} on arbitrary meshes. The method is exemplified for thermal process simulations of unseen parts with inhomogeneous material distribution. Further results show that the model scales well to large and complex meshes, although it is trained on small generic meshes only. 

\acresetall

\end{abstract}





\end{frontmatter}


\section{Introduction}
\label{chap:introduction}

As technology advances and requirements grow, the process of developing technical systems becomes increasingly complex. This is further complicated by shorter development times and a increasing shortage of resources. In particular, components, materials and manufacturing processes must be developed jointly to achieve the technical demands. In order to handle this challenge in a reasonable amount of time, engineers tend towards computer-aided engineering (CAE). Most notably, physics-based simulations have become indispensable in the development of complex technical systems.

Simulations are in general built up on \acp{pde} that describe the considered technical system. Solving \acp{pde} generally requires numerical approximation methods because analytical solutions are not tractable for arbitrary geometries. Therefore, numerical methods such as the \ac{fem} \citep{larsonFiniteElementMethod2013}, the \ac{fvm} \citep{moukalledFiniteVolumeMethod2016c} or the \ac{fdm} \citep{ozisikFiniteDifferenceMethods2017} are frequently applied \citep{venkateshanComputationalMethodsEngineering2023}. These simulations can take hours or even days for complex systems and sufficiently realistic solutions. 
In practice, the computation time even multiplies for parametric studies or iterative optimization \citep{zimmerlingOptimisationManufacturingProcess2022a}. 

To overcome this limitation, surrogate models can be applied. Surrogates are numerically efficient approximations of the "input-output"-relation of an expensive simulation. They are constructed based on a set of a-priori sampled observations. Recently, \ac{ml} methods have gained attention as surrogate models, because they enable huge speed-ups while being fully differentiable and achieving good accuracy. This allows to replace numerical solvers by \ac{ml}-based surrogate models, but their reliability and applicability to unseen scenarios is highly dependent on the model architectures. Many existing models are trained on a specific part and cannot consider new arbitrary part designs, e.g. \citep{longPDENetLearningPDEs2018, pfrommerOptimisationManufacturingProcess2018a}. However, this contradicts many use cases because the part design and its manufacturing process strongly influences the part's function and consequently its quality. This makes it crucial for a learned surrogate model to generalize to arbitrary unseen geometries of complex material systems for changing process conditions. Some approaches seek to replace geometry parameters by pixels- or voxels and analyse them with ~\acp{cnn} \citep{guoConvolutionalNeuralNetworks2016,bhatnagarPredictionAerodynamicFlow2019,zimmerlingOptimisationManufacturingProcess2022a}. This substantially boosts the space of processable geometries but requires to transform input and output to conform to regular grids, which restricts their applicability. In classical numerics, so-called 'meshes' have proven a universal and machine-readable description of geometries/computational domain. Thus, new approaches have been developed, e.g. by message passing \acp{gnn} \citep{pfaffLearningMeshBasedSimulation2020, horiePhysicsEmbeddedNeuralNetworks2022} or transformers \citep{liTransformerPartialDifferential2022}, to tackle this issue and naturally handle simulation data based on arbitrary meshes.

In addition to handling part geometries, neural \acp{pde} solver can also be differentiated in terms of time, e.g. categorized into next-step models and time-dependent neural operator methods \citep{brandstetterMessagePassingNeural2021}. Next-step models predict the solution iteratively, using the solutions of previous time steps as input \citep{sanchez-gonzalezLearningSimulateComplex2020,pfaffLearningMeshBasedSimulation2020,brandstetterMessagePassingNeural2021}. In contrast, time-dependent neural operators learn to predict a solution given the initial condition and its time step, which allows to approximate quickly a solution at future time steps with only one model call \citep{sirignanoDGMDeepLearning2018, raissiPhysicsinformedNeuralNetworks2019}, but limits applicability of the model for out-of-distribution time steps. 
Similar to classical explicit time integration, next-step models can be applied at any starting point and for any simulation duration, making them a flexible option for engineering tasks. However, next-step models depend on their previous output and thus struggle with error accumulation. To overcome this limitation, several strategies have been deployed, such as training noise injection \citep{sanchez-gonzalezLearningSimulateComplex2020, pfaffLearningMeshBasedSimulation2020} or temporal bundling \citep{brandstetterMessagePassingNeural2021}.

While mesh-based ML-techniques show great potential to model arbitrary geometries, their current applications are still data-driven and as such they require vast amounts of training data.
Here, physics-informed training is quickly becoming a popular alternative \citep{sunSurrogateModelingFluid2020, amininiakiPhysicsinformedNeuralNetwork2021,wurthPhysicsinformedNeuralNetworks2023}. Instead of learning the system dynamics from (sufficiently many) supplied samples, physics-informed training seeks to train a model directly on the governing physical equations - usually cast as PDEs. First studies on \acp{pinn} \citep{raissiPhysicsinformedNeuralNetworks2019} use feed-forward neural networks and consider the position on the part as an additional network input, which however restricts the model to learn design-dependent solutions. To alleviate this limitation, several works have transferred the physics-informed training idea to model architectures that perform better on unseen arbitrary part designs, e.g. based on \acp{cnn} \citep{gaoPhyGeoNetPhysicsinformedGeometryadaptive2021,renPhyCRNetPhysicsinformedConvolutionalrecurrent2022,zhaoPhysicsinformedConvolutionalNeural2023}, PointNets \citep{kashefiPhysicsinformedPointNetDeep2022} and \acp{gnn} \citep{gaoPhysicsinformedGraphNeural2022, liIsogeometricAnalysisbasedPhysicsinformed2023}. 
Li et al. \citep{liIsogeometricAnalysisbasedPhysicsinformed2023} propose a hybrid approach of a data-driven loss and an \acl{iga}-based physics-informed loss to train \ac{gnn}-submodels for local approximation, which are input to a purely data-driven \acp{gnn} assembly model. Gao et al. \citep{gaoPhysicsinformedGraphNeural2022} train Chebyshev \acp{gnn} \citep{defferrardConvolutionalNeuralNetworks2016} with an \ac{fem}-based loss function to solve stationary forward and inverse problems for a single given geometry, but the method has not been evaluated on new meshes. Kashefi et al. \citep{kashefiPhysicsinformedPointNetDeep2022} developed a physics-informed training of PointNets based on automatic differentiation and for stationary problems, such as incompressible flows and thermal fields, which was able to handle unseen arbitrary geometries.

The present work introduces \acfp{pimgn}, a physics-informed framework for training \acp{mgn} to reliably simulate non-stationary physics of unseen arbitrary part designs, material property fields and process conditions. \aclp{mgn} are incorporated into the well-known and widely used \ac{fem} to surpass existing \ac{pinn} variants. Unlike prior work, this work trains \acp{mgn} with non-stationary and nonlinear \ac{fem} \acp{pde}, which allows to dispense with time-consuming data generation and even improves generalization and accuracy. Further enhancements are achieved by extending \acp{mgn} with additional methods such as time bundling \citep{brandstetterMessagePassingNeural2021} and global features \citep{battagliaRelationalInductiveBiases2018a}. 

To evaluate the \acp{pimgn}, the effectiveness of the proposed approach is demonstrated for three different time-dependent experiments with different task complexity and learning difficulty. The present work initially compares \acp{pimgn} with data-driven \acp{mgn} \citep{pfaffLearningMeshBasedSimulation2020} and \acp{piggn} \citep{gaoPhysicsinformedGraphNeural2022} and evaluates their improvements over the existing baselines. The subsequent evaluation illustrates the improvement of the trained \acp{pimgn} models over the baselines in predicting non-stationary physics on large unseen parts without retraining. In addition, an ablation study of the \acp{mgn} architecture reveals the contribution of the selected components to achieve reliable performance across all considered tasks.

\section{Methodology}
\label{chap:methodology}
\subsection{Simulating non-stationary physics}
Non-stationary physical phenomena of complex physical systems are usually modelled by time-dependent and generally nonlinear \acp{pde}. Simulating the physics consists of solving these \acp{pde} to obtain the time evolution of the physical quantities of the system. The underlying \acp{pde} differ from application to application, e.g. depending on the physical system and the research objectives.
This work considers parabolic \acp{pde}, described by 

\begin{equation}\label{eqn:pde_parabolic} 
\frac{\partial T}{\partial t} = \alpha \nabla^2 T + q(T) \text{,} ~ \forall \mathbf{x} \in \Omega
\end{equation}
where $T = T(\mathbf{x},t)$ denotes a scalar and time-dependent field of a physical quantity, such as the part temperature in this work, $\alpha = \alpha(\mathbf{x},t)$ the diffusivity of the material, e.g. the thermal diffusivity, and $q(T)$ an in general nonlinear source term. 

An initial-boundary value problem additionally includes boundary conditions on the boundary $\partial \Omega$, e.g. Dirichlet boundary conditions $T = \overline{T}\text{,} ~ \forall \mathbf{x}  \in  \Gamma \subset \partial \Omega$  or Neumann boundary conditions $\nabla T \cdot \mathbf{n} = h_{\mathcal{N}}\text{,} ~ \forall \mathbf{x}  \in \partial \Omega_{\mathcal{N}} = \partial \Omega \setminus \Gamma $, where $\mathbf{n}$ denotes the normal vector on the boundary $\partial \Omega_{\mathcal{N}}$, and an initial condition $T(\mathbf{x},t=0) = T^0$.
The solution $T$ of the initial-boundary value problem can be calculated with the \ac{fem}, which will be briefly revisited with the pertinent equations as necessary to outline the concept of PI-MGNs in the following.

\subsection{The finite element method} \label{subsec:fem}

In the \ac{fem}, the weak formulation of the \acp{pde} is considered, which can be obtained by multiplying the strong \ac{pde} of \Cref{eqn:pde_parabolic} with a test function $\varphi \in H^1_0$ and integration by parts \citep{larsonFiniteElementMethod2013} as
\begin{equation} \label{eqn:weak_pde} 
  \int_{\Omega} \frac{\partial T}{\partial t} \varphi + \alpha \nabla T \cdot \nabla \varphi -q(T)\varphi ~\text{d} V - \int_{\partial \Omega_{\mathcal{N}}}  \alpha h_{\mathcal{N}}\varphi  ~\text{d} A = 0\text{,}
\end{equation}
or, more generally, 
\begin{equation} \label{eqn:general_weak_pde} 
  \Phi\left(\frac{\partial T}{\partial t},T,\varphi\right) = 0\text{.}
\end{equation}
 Here, $H^1_0$ denotes a Sobolev space in which $\varphi=0$ on $\partial \Omega$ and $\varphi$ as well as its first weak derivative is square integrable \citep{larsonFiniteElementMethod2013}. 
The Galerkin \ac{fem} uses the same basis functions, e.g. piecewise linear polynomials in this work, to approximate the solution \mbox{$T(\textbf{x}, t) = \sum_{v=1}^{N_\varphi} T_v(t) \phi_v(\textbf{x})$} and the test function $\varphi(\textbf{x}) = \sum_{v=1}^{N_\varphi} \phi_v(\textbf{x})$, i.e., 

\begin{equation*}
\label{eqn:weak_pde_linear_basis_function}
\sum_{v=1}^{N_\varphi} \Phi\left(\frac{\partial T_v}{\partial t},T_v,\phi_v, \phi_m\right) = 0, ~\forall m = 1,...,N_{\varphi}\text{.}
\end{equation*}

Then, the weak form is discretized in time. This work uses an implicit Euler method with a constant time step $\Delta t$ as
\begin{equation*}
\begin{aligned}
\sum_{v=1}^{N_\varphi} \Phi\left(\frac{\partial T_v}{\partial t},T_v,\phi_v, \phi_m\right) &\approx \sum_{v=1}^{N_\varphi} \Phi\left((T_v^{n+1} - T_v^{n})/\Delta t,T_v^{n+1},\phi_v, \phi_m\right) \\
&\approx \sum_{v=1}^{N_\varphi} \Phi\left(T_v^{n+1}, T_v^{n},\phi_v, \phi_m\right)\text{,}
\end{aligned}
\end{equation*}
resulting in $N_t$ equations for the solution $T_v^{n}  \in \{T_v^{0},...T_v^{N_t}\}$ and the time steps $t^n \in \{t^0,...t^{N_t}\}$ with $n=0,..,N_t$. 
Subsequently, the integral over the domain $\Omega$ is separated into element wise integrals 

\begin{equation}
\label{eqn:discretisized_weak} 
\sum_{e=1}^{N_{\text{e}}} \sum_{v=1}^{N_\varphi}  \Phi_{e}\left(T_v^{n+1}, T_v^{n},\phi_v, \phi_m\right)= 0, ~\forall m = 1,...,N_{\varphi}
\end{equation}

and the element integrals $\Phi_{e}$ are approximated as a second-order accurate quadrature in this work.

Finally, the discretized solution $T_v^{n+1}$ can be obtained by starting with the initial condition $T_v^0$ and iteratively solving \Cref{eqn:discretisized_weak} for the next time step $T_v^{n+1}$ for all time steps $t^n \in \{t^0,...t^{N_t}\}$. The applied backward Euler time discretization provides stability for large time steps \citep{larsonFiniteElementMethod2013}. This is necessary for fast simulations of stiff \acp{pde}, where the solution changes strongly at different time scales and explicit methods are in consequence only stable for small time steps \citep{larsonFiniteElementMethod2013}. However, solving for one time step of nonlinear \acp{pde} with an implicit numerical solver requires multiple inner solver iterations, which can be very time-consuming \citep{larsonFiniteElementMethod2013}. Consequently, stiff and nonlinear PDEs are particularly difficult to solve numerically and would benefit in particular from a more efficient solution approach.

Applying a given Dirichlet boundary condition simplifies the problem by pre-determining the coefficients $T_{v,\Gamma}^{n+1}$ associated with the basis functions that are $1$ at the Dirichlet boundary $x\in\Gamma$. This simplification reduces the number of unknown coefficients from $N_\varphi$ to $N_{\varphi,\mathcal{F}}$.
To prevent an overdetermined system of \acp{pde}, \Cref{eqn:discretisized_weak} is solved only for the $m = 1,..,N_{\varphi,\mathcal{F}}$ test functions \mbox{$\phi_{m,\mathcal{F}} \in \{\phi_m(\mathbf{x}): \phi_m(\mathbf{x}) \neq 1, \forall \mathbf{x} \in \Gamma \}$}, which are not $1$ on the Dirichlet boundary $\Gamma$. Then, it is sufficient to predict the solution $T_{v,\mathcal{F}}^{n+1}$ of the nodes that are not on the Dirichlet boundary $\Gamma$. Finally, the solution is obtained by enforcing the solution on the Dirichlet boundary as $T_{v}^{n+1} = T_{v,\mathcal{F}}^{n+1} + T_{v,\Gamma}^{n+1}$, where  $T_{v,\Gamma}^{n+1}$ is non-zero on the nodes of $\Gamma$ and $T_{v,\mathcal{F}}^{n+1}$ is non-zero on $\Omega \setminus \Gamma$.

\subsection{\textsc{MeshGraphNets}} \label{subsec:gns}
The solving process of the \ac{fem} can be speed up by learned \ac{ml} simulators.
Here, an \ac{mgn} based on a \ac{mpn} with global features is applied. \\
The \ac{mpn} processes graphs $\mathcal{G} = (\mathcal{V},\mathcal{E},\mathbf{X}_{\mathcal{V}},\mathbf{X}_{\mathcal{E}},\mathbf{g})$ by updating the node features $\mathbf{X}_{\mathcal{V}}$ and edge features $\mathbf{X}_{\mathcal{E}}$ of all nodes $v \in \mathcal{V}$ and edges $\varepsilon \in \mathcal{E}$, as well as the global features $\mathbf{g}$ in $L$ consecutive message passing steps. The $l$-th message passing step for updating the node feature $\mathbf{x}^{l}_v$, edge feature $\mathbf{x}^{l}_\varepsilon$ and global feature $\mathbf{g}^{l}$ is defined as 

\begin{equation} \label{eqn:mpn}
\begin{aligned}
\mathbf{x}^{l+1}_{\varepsilon} &= f^{l}_{\mathcal{E}}(\mathbf{x}^{l}_v, \mathbf{x}^{l}_u, \mathbf{x}^{l}_{\varepsilon}, \mathbf{g}^l), \textrm{ with } \varepsilon = (u, v) \in \mathcal{E}~\text{and}~ u,v \in \mathcal{V} \text{,} \\
\mathbf{x}^{l+1}_{v} &= f^{l}_{\mathcal{V}}(\mathbf{x}^{l}_{v}, \bigoplus_{\varepsilon=(v,u)\in \mathcal{E}} \mathbf{x}^{l+1}_{\varepsilon}, \mathbf{g}^l)\text{,}\quad \text{~and~} \\
\mathbf{g}^{l+1} &=
f^{l}_{\mathbf{g}}(\bigoplus_{v\in\mathcal{V}}\mathbf{x}^{l+1}_{v}, \bigoplus_{\varepsilon\in \mathcal{E}} \mathbf{x}^{l+1}_{\varepsilon}, \mathbf{g}^l)\text{.}
\end{aligned}
\end{equation}

The nonlinear learned functions $f^l_.$ are usually implemented as \acp{mlp}, and $\bigoplus$ denotes a permutation-invariant aggregation, such as a sum, mean or max operator \citep{battagliaRelationalInductiveBiases2018a,pfaffLearningMeshBasedSimulation2020,freymuthSwarmReinforcementLearning2023}.

The \ac{mpn} outputs learned quantities $x^L_v$ per node.
These values are then combined with 1D \acp{cnn} to reduce error accumulation by predicting a solution over multiple time steps, which is known as temporal bundling~\citep{brandstetterMessagePassingNeural2021}.
In particular, each time step $t^n$ of a problem $p$ induces a directed graph $\mathcal{G}^n$ with self-loops that contains \ac{pde}-specific node features $\mathbf{X}_{\mathcal{V}}^n$, edge features $\mathbf{X}_{\mathcal{E}}^n$ and global features $\mathbf{g}^n$. The nodes $v^n$ of the graph $\mathcal{G}^n$ correspond to mesh nodes at time step $t^n$ and the edges $\varepsilon^n$ to the edges of the mesh. To predict the temporal bundle \mbox{$\tilde{T}_v^{n+\tilde{n}}$} of solutions for the next $\tilde{n} = 1, ..., N_{\text{TB}}$ time steps, the features of the graph are encoded by individual MLPs, and then subsequently processed by the \ac{mpn} and the \ac{cnn} decoder. The resulting bundle can be used to assemble the next graph $\mathcal{G}^{n+N_{\text{TB}}}$ to be processed by the \ac{mgn}. This procedure is repeated until the desired time step $t^{N_t}$ is reached. This methodology can be applied to batched graphs $\mathcal{G}_\text{b}^n$ to better utilize accelerated hardware.  \\

In this work, a one-hot vector $n_\text{t}$ \citep{pfaffLearningMeshBasedSimulation2020} is added to the graph as a node feature, which encodes the node type, categorized into Dirichlet nodes, Neumann nodes and inner nodes. The relative distance vector between the nodes of the edge $\mathbf{x}_{vu}  = \mathbf{x}_v - \mathbf{x}_u$  as well as the Euclidean distance $|\mathbf{x}_{vu}|$  are added as edge features. In addition, the difference of the solution $T_{vu} = T_v^{n}-T_u^{n}$, respectively, the approximation $\tilde{T}_{vu} = \tilde{T}_v^{n}-\tilde{T}_u^{n}$ at the current time step $t^n$ between the nodes of an edge $\varepsilon = (u,v)$ are added. Additional node, edge or global features may be added individually in the specific experiments. An overview is given in the Appendix in \cref{table:graph_feauters}.

\subsection{Physics-informed \textsc{MeshGraphNets}}
MGNs as described in \Cref{subsec:gns} are commonly trained in a data-driven fashion to learn to solve \Cref{eqn:discretisized_weak} for unseen tasks, e.g new part designs. This training paradigm requires simulation data that is created using numerical solvers for multiple training tasks, which often proves time- and resource-intensive.
Therefore, the present work trains \acp{mgn} with a physics-informed approach that builds on the \ac{fem} formulation of \Cref{subsec:fem}, learning to predict the non-stationary physics for unseen part designs, materials and process settings.

\paragraph{FEM-based loss} A correct solution of \Cref{eqn:discretisized_weak} yields zero on the left side of the equation for all $N_{\varphi,\mathcal{F}}$ equations, and a non-zero value otherwise. Given a solution of the current time step $T_v^{n}$, the left side of \Cref{eqn:discretisized_weak} can be interpreted as an error function for the approximation $\tilde{T}_v^{n+1}$ of the \ac{mgn} at the time step $n+1$.
For the $m$-th test function, the error is thus given as 
\begin{equation}
\label{eqn:pi_error} 
\epsilon^{n+1}_{\text{FEM},m} = \sum_{e=1}^{N_{\text{e}}} \sum_{v=1}^{N_\varphi}  \Phi_{e}\left(\tilde{T}_v^{n+1}, T_v^{n},\phi_v, \phi_m\right), ~\forall m = 1,...,N_{\varphi,\mathcal{F}} \text{,}
\end{equation}
which is then similar to the loss in \citep{gaoPhysicsinformedGraphNeural2022} for a steady \ac{pde}.
Naively evaluating this formulation requires $\sum_{v=1}^{N_\varphi} \Phi_{e}\left(\tilde{T}_v^{n+1}, T_v^{n},\phi_v, \phi_m\right)$ to be evaluated in $N_{\text{e}}$ elements for all $N_{\varphi,\mathcal{F}}$ test functions at each time step $n$, which is quite expensive. 
However, the complexity of the calculation can be simplified. The piecewise polynomials $\phi_{m}$ in $\Phi_{e,km}$ are only non-zero for the elements they are connected to and, in consequence, $\Phi_{e}$ is only non-zero for these elements.
This simplification allows to split up the calculation of $\epsilon^{n+1}_{\text{FEM}}$ in two steps.
First, an element-wise error is calculated for each of the test functions $\phi_{m}$, which are non-zero in the element, otherwise the error is set to zero. Formally, this amounts to, 

\begin{equation} \label{eqn:element_wise_loss}
 \epsilon_{\text{FEM},em}^{n+1}  = 
\left\{
\begin{aligned}
&\sum_{v=1}^{N_\varphi}  \Phi_e\left(\tilde{T}_v^{n+1}, T_v^{n},\phi_v\text{,} \phi_m\right), ~&\phi_{m} \neq 0 \\
&0\text{,} ~&\phi_{m} = 0
\end{aligned}
\right.
\end{equation}
for all $m = 1,\dots,N_{\varphi,\mathcal{F}}$ and $e = 1,\dots,N_{\text{e}}$. 

The total error $\epsilon^{n+1}_{\text{FEM},m}$ of the $N_{\varphi,\mathcal{F}}$ test functions at the time step $n$ yields
\begin{equation} \label{eqn:test_function_loss}
  \epsilon^{n+1}_{\text{FEM},m} =  \sum_{e=1}^{N_{\text{e}}}  \epsilon_{\text{FEM},em}^{n+1} ~\text{,} ~\forall m = 1,...,N_{\varphi,\mathcal{F}} \text{.}
\end{equation}
Finally, the loss is defined as the \ac{mse} of this error 
\begin{equation} \label{eqn:pi_loss}
  \mathcal{L}^{n}_{\text{FEM}} = \frac{1}{N_{\text{TB}} N_{\varphi,\mathcal{F}}} \sum_{\tilde{n}=1}^{ N_{\text{TB}}} \sum_{m=1}^{N_{\varphi,\mathcal{F}}}   \left( \epsilon^{n+\tilde{n}}_{\text{FEM},m} \right)^2 ~\text{,} 
\end{equation}
for all time steps $\tilde{n} = 1,...,N_{\text{TB}}$ of the temporal bundle, which
enables a physics-informed training of the \ac{mgn}.

\begin{figure}[!htb]
    \centering
    \includegraphics[]{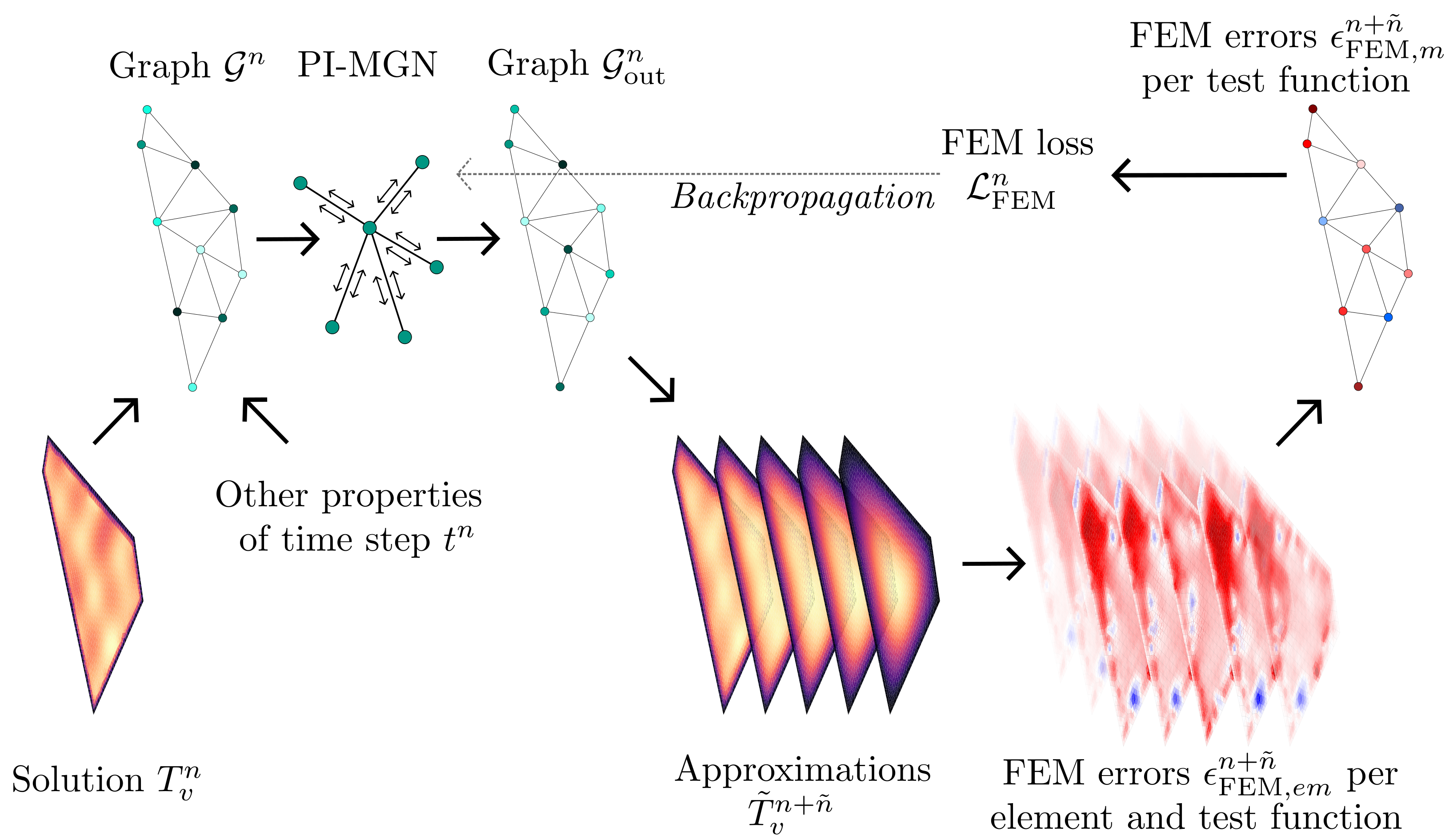}
    \caption{Given a solution $T_v^n$ and additional properties of the time step $t^n$, a graph $G^n$ is assembled and passed to the \acp{pimgn}, which output a graph $G^n_{\text{out}}$. This graph contains the approximations $\tilde{T}_v^{n+\tilde{n}}$ of the next time steps $t^{n+\tilde{n}}$, where $\tilde{n} = 1, ..., N_{\text{TB}}$. Using the \ac{fem}, an element- and test function-wise error is calculated and summed up to a test function-wise error. Finally, the loss is calculated as the \ac{mse} of all errors to update the \ac{pimgn} using backpropagation.}
    \label{fig:optimization_step}
\end{figure}

\paragraph{\acp{pimgn} training} The physics-informed training is divided into multiple epochs, with each epoch containing several optimization loops with multiple optimization steps. One optimization step (cf. \Cref{fig:optimization_step}) consists of: The \ac{pimgn} processes a graph $\mathcal{G}^n$ and outputs the solution $\tilde{T}_v^{n+\tilde{n}}$ for the next temporal bundle, containing the solutions of the next $\tilde{n} = 1, ... , N_{\text{TB}}$ time steps. The element-wise errors for the $N_{\text{TB}}$ time steps are evaluated in parallel with \Cref{eqn:element_wise_loss}, using the initial solution $T_v^{(n)}$ and the approximations $\tilde{T}_v^{n+\tilde{n}}$. Then, the error per test function is calculated with \Cref{eqn:test_function_loss}. For each of the $N_{\text{TB}}$ time steps, $N_{\varphi,\mathcal{F}}$ errors are obtained, resulting in a total amount of $N_{\text{TB}} N_{\varphi,\mathcal{F}}$ errors for the $N_{\text{TB}} N_{\varphi,\mathcal{F}}$ approximations of the \acp{pimgn}. The \ac{fem} loss $\mathcal{L}_{\text{FEM}}^{n}$ for the graph input $\mathcal{G}^n$ is calculated as the \ac{mse} of $N_{\text{TB}} N_{\varphi,\mathcal{F}}$ all errors. Finally, the loss $\mathcal{L}_{\text{FEM}}^{n}$ is used to update the \ac{pimgn} parameters using backpropagation through the \ac{fem} equations and a gradient-based optimizer \citep{kingma2014adam}. 

One optimization loop starts with the initial condition at $t^0$ of a problem and assembles a graph $\mathcal{G}^0$ to update the \ac{pimgn}. After this step, the time step $t^{N_{\text{TB}}}$ can be initialized to perform another optimization step. This procedure is repeated until the last time step $t^{N_{\text{T}}}$ is reached. 
Then, the optimization loop processes the next problem. The epoch is finished when the optimization loop has been repeated for all considered training problems $p$. The full training procedure consists of a given number of epochs.

It is important to note that the method is fully applicable to nonlinear equations without restrictions. The loss can also be calculated directly and in parallel for multiple time steps without the need for inner solving iterations. Additionally, since the FEM naturally integrates the Neumann boundary conditions into the \acp{pde} and the Dirichlet boundary conditions are strictly enforced in this work, the initial-boundary-value problem has only to be solved for \cref{eqn:pi_loss}. This directly avoids competing training losses and the resulting training difficulties \citep{bischofMultiObjectiveLossBalancing2023}.

\paragraph{Noise injection in physics-informed training} As the output of an \ac{mgn} at the time step $n$  is part of the next input graph $\mathcal{G}^{n+N_{\text{TB}}}$, next-step models struggle with error accumulation. Noise injection proves to be an effective strategy for \ac{mgn} \citep{pfaffLearningMeshBasedSimulation2020} to reduce this error accumulation. The basic idea is to add an artificial error to the input so that the \ac{mgn} must detect and minimize the error. Here, noise injection is introduced for physics-informed training. Using Gaussian noise $\epsilon\sim\mathcal{N}(0,\mathbf{I}\sigma)$, a noisy solution difference $(T_v^{n}+\epsilon_v)-(T_u^{n}+\epsilon_u)$ respectively approximation $(\tilde{T}_v^{n}+\epsilon_v)-(\tilde{T}_u^{n}+\epsilon_u)$ is used as a substitute to the exact solution difference during training. However, the noiseless solution $T_v^{n-1}$ respectively approximation $\tilde{T}_v^{n-1}$ is still inserted into \Cref{eqn:pi_loss}, enabling the \ac{pimgn} to learn to output the correct solution of \Cref{eqn:pi_loss} for a noisy input. As in \citep{pfaffLearningMeshBasedSimulation2020}, the noise is only added during training and not at inference.

\section{Experiments}
\label{chap:experiments}
\subsection{Training on small meshes}
\label{subsec:exp_small_meshes}
\begin{figure}[!htb]
    \centering
    \includegraphics[]{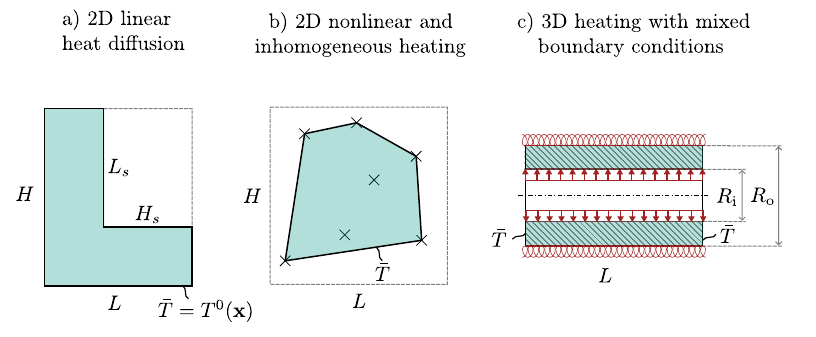}
    \caption{Schematic overview of the experiments. Figure a) shows an example of the L-shaped domain of a 2D linear heat diffusion experiment and Figure b) the initialization of an convex polygon with $7$ random points of the nonlinear experiment. Figure c) depicts a hollow cylinder of the 3D experiment with a heating boundary condition inside the cylinder (red arrows) and an adiabatic boundary condition at the outer surface (red coils).}
    \label{fig:experiments}
\end{figure}

The \acp{pimgn} are evaluated on three different experiments (cf. \cref{fig:experiments}), where the models face different numerical and learning challenges:

\begin{itemize}
    \item Firstly, a 2D linear heat diffusion task on randomly varying L-shaped domains is considered. The model learns to handle unseen meshed geometries and different process conditions in the form of random initial conditions. 
    \item The second experiment contains a \ac{pde} with a nonlinear, time- and material-dependent heat source and considers random 2D convex polygons with random material distribution. The model has to learn the influence of random arbitrary material fields on the solution of a nonlinear \ac{pde}.
    \item Finally, the method is applied to various 3D meshes with mixed boundary conditions, in particular hollow cylinders that are heated over time due to a Neumann boundary condition.
\end{itemize}
\ref{chap:appendix_A} describes the three experiments in more detail.

\paragraph{Training} In the following, a problem $p$ contains a concrete specification of the experiments. In particular, it includes the specification of all material parameters, such as the diffusivity $\alpha$, the nonlinear source term $q(T)$ in \Cref{eqn:pde_parabolic}, a geometry $\Omega$, the initial condition $T^0$, the Dirichlet $\overline{T}\text{,} ~ \forall \mathbf{x} \in  \Gamma$ and Neumann boundary conditions $h_{\mathcal{N}}\text{,} ~ \forall \mathbf{x}  \in \partial \Omega_{\mathcal{N}}$. All experiments consider a total number of $100$ randomly generated problems. $75$ of these problems are selected randomly for training, while the other $25$ are used for validation. Training is repeated over $5$ random seeds, and each training consists of $500$ epochs with mini-batches comprised of $2$ graphs. The Adam~\citep{kingma2014adam} optimizer with an exponential learning rate decay from $\num{1e-3}$ to $\num{1e-5}$ is used.
The gradient norm is clipped to $1$.

\paragraph{\ac{pimgn}} All \acp{pimgn} consist of $12$ message passing steps and output the predicted solution of the next $20$ simulation steps. Each \ac{mlp} consists of $2$ layers of $128$ neurons and uses a \ac{relu} activation function between layers and a layer norm~\citep{ba2016layer} at the end. The input vector of node features, edge features and global features respectively is linearly encoded to a latent vector of the hidden dimension of $128$. The latent vector of size $128$ is decoded by a \ac{cnn} to output a vector of length $20$, containing the solution approximation of $20$ time steps. The setup of the \ac{cnn} follows previous work \citep{brandstetterMessagePassingNeural2021}, consisting of $2$ layers with $1$ input channel, $8$ hidden channels and $1$ output channel, a kernel stride of $15$ respectively $10$ and a stride of $4$ respectively $1$. 

\paragraph{Simulation and Evaluation} All problems are evaluated for a total amount of $100$ time steps, starting from $t^0 = \SI{0}{s}$ and progress until $t^{N_t}= \SI{1}{s}$ is reached with a fixed time step $\Delta t = \SI{0.01}{s}$. For evaluation, FEM-solutions from the open-source package \textit{scikit-fem} \citep{gustafssonScikitfemPythonPackage2020a} serve as a ground truth reference.
The \ac{fem}-solutions solve the same equations and time steps as the \acp{pimgn} for comparability. These FEM-solutions are in no way used for \ac{pimgn} training. They only serve as ground truth to assess the quality of \ac{pimgn} solutions after training. 
A normalized $L_2$ error
\begin{equation}
\epsilon_{L_2,\text{norm}} = \frac{\left(   \sum_{n=1}^{N_{t}} \sum_{v=1}^{N_\varphi} \left| \tilde{T}_v^n - T_v^n \right|^2 \right)^{1/2}}{\left(  \sum_{n=1}^{N_{t}} \sum_{v=1}^{N_\varphi}  \left| T_v^n \right|^2 \right)^{1/2}}
\end{equation}
is used to measure the deviation of the \ac{pimgn} prediction to the ground truth.
For visualizing error distributions, a relative metric

\begin{equation}
\epsilon_{\text{rel}}^{n} = \frac{\tilde{T}_v^n - T_v^n }{\underset{v,n}{\max}( T_v^n)-\underset{v,n}{\min}( T_v^n)} 
\end{equation}
is applied.

\paragraph{Baselines} The \acp{pimgn} are compared to \acp{ddmgn} \citep{pfaffLearningMeshBasedSimulation2020} and to \acp{piggn} from \citep{gaoPhysicsinformedGraphNeural2022}, which train \ac{gccn} with the \ac{fem} for solving forward and inverse stationary problems of a single geometry. 

For the \acp{ddmgn} baseline, the ground truth \ac{fem} data $T_v^{n}$ is integrated into the training process, which amounts to $75$ simulations per experiment and per repetition for training and $25$ for validation. Therefore, the loss function in \Cref{eqn:pi_loss} is replaced by a data-based \ac{mse} loss

\begin{equation}
\mathcal{L}_{\text{data}}^{n} = \frac{1}{N_{\text{TB}} N_{\varphi,\mathcal{F}}} \sum_{\tilde{n}=1}^{N_{\text{TB}}} \sum_{v=1}^{N_{\varphi,\mathcal{F}}} \left| \tilde{T}_v^{n+\tilde{n}} - T_v^{n+ \tilde{n}} \right|^2 \text{.}
\end{equation}
Analogous to the physics-informed loss $\mathcal{L}_{\text{FEM}}^{n}$, the data loss measures the deviation of all $N_{\text{TB}}$ approximations of the \ac{mgn} for an input graph $\mathcal{G}^n$ at the time step $n$. Since the ground truth is now available during training, the noisy ground truth difference $(T_v^{n}+\epsilon_v)-(T_u^{n}+\epsilon_u)$ is added as an edge feature instead of the noisy approximation difference $(\tilde{T}_v^{n}+\epsilon_v)-(\tilde{T}_u^{n}+\epsilon_u)$, as in \citep{pfaffLearningMeshBasedSimulation2020}. Apart from these two changes, everything remains the same for comparability, which includes all integrated extensions such as time bundling and global features. Hence, the \ac{ddmgn} and \ac{pimgn} approaches differ only in their training and work identically during evaluation. 

For the physics-informed baseline \ac{piggn}, the model is integrated into the \acp{pimgn} framework. Therefore, the \ac{gccn} of \citep{gaoPhysicsinformedGraphNeural2022}  replaces the \ac{mpn} of this work and the absolute positions and the solution are added as node features as in \citep{gaoPhysicsinformedGraphNeural2022}, because the \ac{gccn} cannot consider edge features. The input size of the model is set to the experiment-specific numbers of node features and the output size to the time-bundling size $N_{\text{TB}} = 20$. All other components of the \ac{gccn} remain the same as in \citep{gaoPhysicsinformedGraphNeural2022}, while all setups, such as the training setup, are consistent with the \ac{pimgn} setup.

\subsection{Generalization to large unseen meshes}
\label{subsec:exp_generalization}

In addition, the present work examines the ability of the \acp{pimgn} to generalize to large unseen meshes. Therefore, the pre-trained \acp{pimgn} and baseline models are applied on four unseen large mesh problems. The first two problems investigate the performance of the trained \acp{pimgn} to predict the 2D linear heat diffusion for large corrugated sheets consisting of $10$ respectively $100$ components. Subsequently, the models of the 2D nonlinear heating are applied on a large grid structure. The last problem considers the \acp{pimgn} of the 3D heatup with mixed boundary conditions applied on a long 3D hollow cylinder. \ref{subsec:apx_generalization} contains a detailed description of the large mesh problems.

\paragraph{Ablation studies}
Ablation studies investigate the effectiveness of the individual components of the \ac{pimgn} architecture. 
In particular, the time bundling \ac{cnn} decoder is compared to an \ac{mlp} decoder and the sensitivity of the accuracy with respect to the time bundling size $N_{\text{TB}}$ is investigated.
Further, the relative positional encoding is compared to an absolute positional encoding and the effectiveness of the global features and the training noise is investigated.
Finally, the best-performing variants from the ablation study will be applied to the large mesh experiments to evaluate their generalization ability to generalize to large parts.

\paragraph{Speed comparison to the \ac{fem}} Finally, to evaluate the speed of the \ac{pimgn}, the computation time is compared for the 2D nonlinear and inhomogeneous heating, as those are the most challenging for the \ac{fem} due to the nonlinearity.

\section{Results}
\label{chap:results}
\subsection{Training on small meshes}
\label{subsec:res_small_meshes}
\begin{figure}[H]
    \centering
    \includegraphics[]{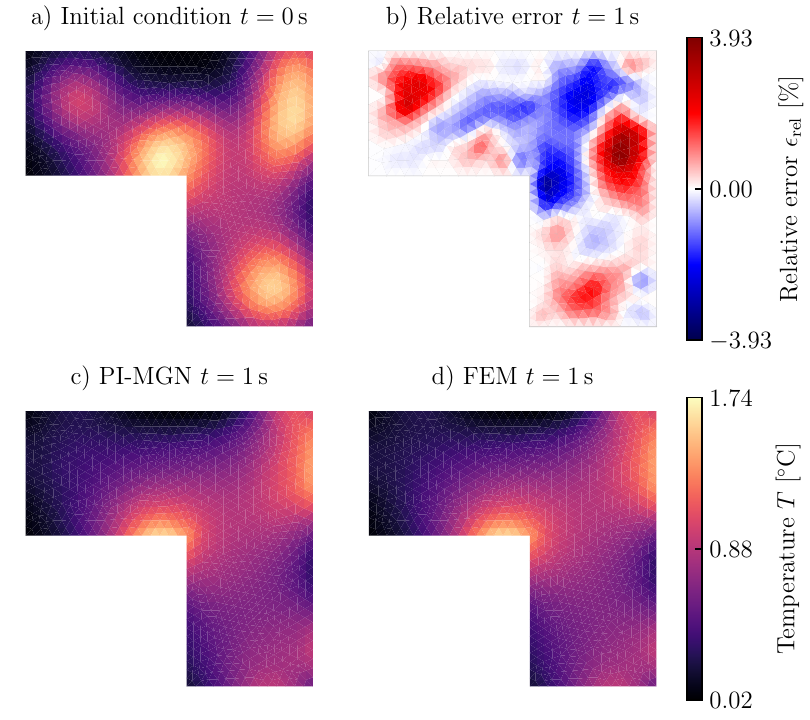}
    \caption{Comparison of the \ac{pimgn} approximation to the \ac{fem} solution at the last time step $t=\SI{1}{s}$ for an unseen problem of the 2D linear experiment of \Cref{subsec:exp_small_meshes}. \Cref{fig:UHDL} a) shows the randomly sampled initial condition on a new L-shaped mesh geometry. The relative error of the \ac{pimgn} in \Cref{fig:UHDL} b) measures the deviation of the \ac{pimgn} approximation in \Cref{fig:UHDL} c) from the ground truth in \Cref{fig:UHDL} d).}
    \label{fig:UHDL}
\end{figure}

\begin{figure}[H]
    \centering
    \includegraphics[width=\textwidth]{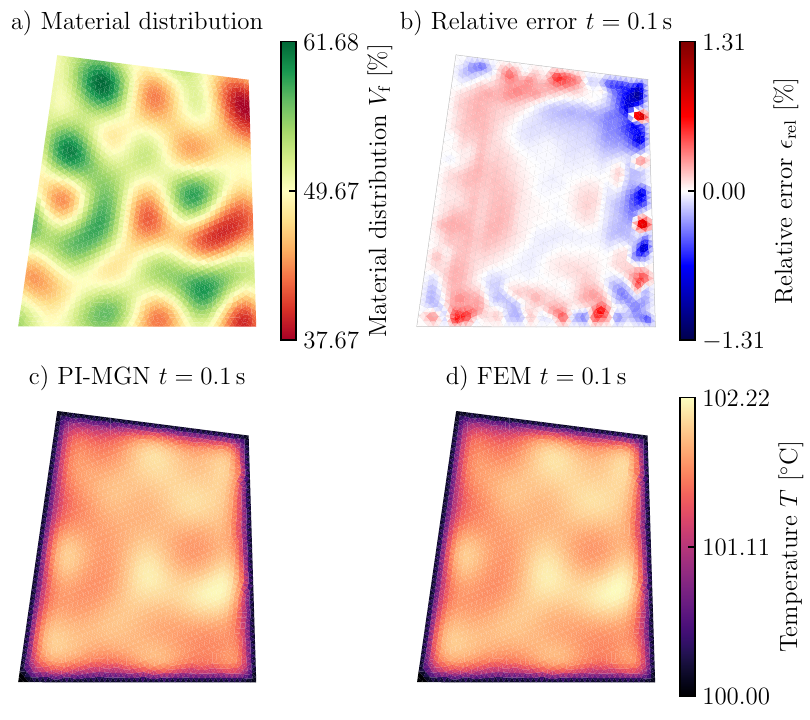}
    \caption{ The \ac{pimgn} solution is compared to the \ac{fem} for an nonlinear problem of the experiment of \Cref{subsec:exp_small_meshes} after $t=\SI{0.1}{s}$. The geometry and material distribution (cf. \Cref{fig:UHDNL} a)) of the part was unseen during training. \Cref{fig:UHDNL} c) and \Cref{fig:UHDNL} d) show the approximation of the \ac{pimgn} respectively the solution of the \ac{fem}. \Cref{fig:UHDNL} d) depicts the relative error between both solutions. }
    \label{fig:UHDNL}
\end{figure}

\begin{figure}[H]
    \centering
    \includegraphics[]{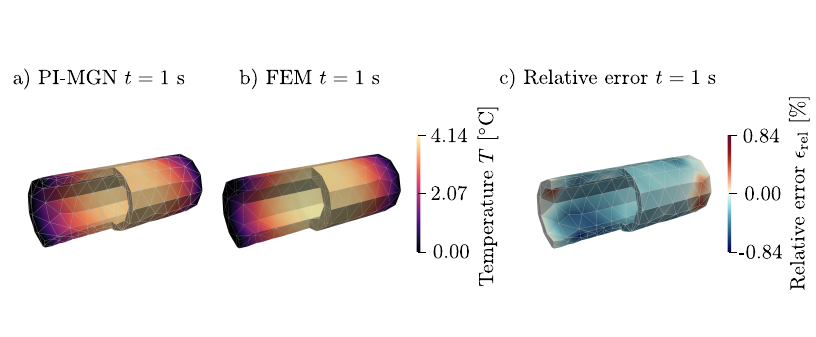}
    \caption{The 3D experiment of \Cref{subsec:exp_small_meshes} with mixed boundary conditions. \Cref{fig:UHDL3D} a) shows the solution of the \ac{pimgn} at the last time step $t=\SI{1}{s}$ and \Cref{fig:UHDL3D} b) the ground truth \ac{fem} solution. \Cref{fig:UHDL3D} c) depicts the relative error of \Cref{fig:UHDL3D} a) compared to \Cref{fig:UHDL3D} b). A quarter of the 3D hollow cylinders is cut off for visualization in \Cref{fig:UHDL3D} a), b) and c).}
    \label{fig:UHDL3D}
\end{figure}

\paragraph{Qualitative comparison to the \ac{fem}} For a qualitative comparison of the accuracy and capability of the trained models, the solutions provided by \ac{pimgn} are compared visually to the ground truth \ac{fem} on unseen part designs, material distributions, and process settings in \Cref{fig:UHDL,fig:UHDNL,fig:UHDL3D}. The results indicate that \ac{pimgn} can produce accurate predictions that are visually similar to \ac{fem} solutions.

More precisely, \Cref{fig:UHDL} a) shows a challenging example of the 2D linear heat diffusion experiment of \Cref{subsec:exp_small_meshes}, which 
contains a random initial condition with 4 hot spots on an unseen part design. 
It can be seen that the \ac{pimgn} has learned to accurately predict the final solution at the last time step $t=\SI{1}{s}$. \Cref{fig:UHDL} b) depicts the relative error $\epsilon_{\text{rel}}$.

\Cref{fig:UHDNL} shows a nonlinear heating problem from \Cref{subsec:exp_small_meshes} with a challenging material distribution on a randomly selected seed to demonstrate that the \acp{pimgn} learn the influence of nonlinear material fields. \Cref{fig:UHDNL} a) shows the unseen material distribution $V_{\text{f}}$ on the unseen part. \Cref{fig:UHDNL} d) depicts the ground truth solution, showing that spots with low values of the material property result in a locally stronger heating. \Cref{fig:UHDNL} c) demonstrate that the \ac{pimgn} precisely reproduces these solution. \Cref{fig:UHDNL} b) depicts the relative error between both simulations, showing that comparatively high deviations mostly appear close to the boundary.

Finally, the qualitative performance of the \ac{pimgn} is compared to the \ac{fem} for an unseen and randomly selected instance of the 3D heating task with mixed boundary conditions of \Cref{subsec:exp_small_meshes}. \Cref{fig:UHDL3D} a) shows the approximation of the \ac{pimgn}, \Cref{fig:UHDL3D} b) the ground truth and \Cref{fig:UHDL3D} c) the relative error at the last time step $t=\SI{1}{s}$ of the simulation. A quarter of the unseen 3D hollow cylinders is cut off for visualization, providing a view of the solution inside the cylinder. As before, the \ac{pimgn} and \ac{fem} solutions are in good agreement. Only a slight underestimation of the \ac{pimgn} is visible. 


The results show that the \ac{pimgn} can handle non-stationary and nonlinear \acp{pde} in 2D and 3D, enabling visibly precise prediction for the unseen part designs, inhomogeneous material fields under new process conditions.

\begin{figure}[!htb]
    \centering
    \includegraphics[]{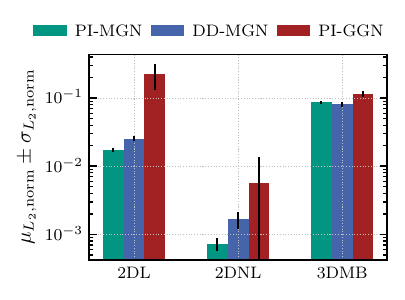}
    \caption{Comparison of the \acp{pimgn} to \acp{ddmgn} and to the physics-informed baseline \ac{piggn}. \Cref{fig:comp_base} shows the mean $\mu_{L_2,\text{norm}}$ (bar) and the standard deviation $\sigma_{L_2,\text{norm}}$ (black line onn bar) of the normalized $L_2$ error $\epsilon_{L_2,\text{norm}}$, calculated for five training repetitions per model and experiment. Each bar group contains the results of the \acp{pimgn}, \acp{ddmgn} and \acp{piggn} for one of the three experiments of \Cref{subsec:exp_small_meshes}.}
    \label{fig:comp_base}
\end{figure}

\paragraph{Comparison to baseline models} \Cref{fig:comp_base} shows the performance of the \acp{pimgn} relative to the baselines \acp{ddmgn} and \acp{piggn} for all three experiments: The 2D linear heat diffusion ('\textit{2DL}'), the 2D nonlinear heating ('\textit{2DNL}') and the 3D heating with mixed boundary conditions ('\textit{3DMB}').  The mean $\mu_{L_2,\text{norm}}$ and the standard deviation $\sigma_{L_2,\text{norm}}$ of the normalized $L_2$ error $\epsilon_{L_2,\text{norm}}$ over five training repetitions is displayed for each method, calculated for 25 unseen evaluation problems. The comparability of absolute values between experiments is limited as all values are normalized by their reference solution. For example, the mean temperature in the 2D nonlinear experiment is about $\SI{100}{\degreeCelsius}$, while it is $\SI{0}{\degreeCelsius}$ in the others.

The \ac{pimgn} outperforms the \ac{ddmgn} in two out of three experiments. The mean error $\mu_{L_2,\text{norm}}$ of the data-driven training is almost $\SI{50}{\%}$ higher for the 2D linear heatup experiment and approximately doubles for the 2D nonlinear and inhomogeneous heating. 
In contrast, the mean error of \acp{pimgn} is only slightly higher than that of \acp{ddmgn} for the 3D heatup task. 
Here, the standard deviations overlap. Furthermore, the standard deviation $\sigma_{L_2,\text{norm}}$ of the data-driven approach is larger in all three tasks, with an increase of up to $\SI{200}{\%}$. 
For this 3D experiment, the physics does not vary as much across problems as in the other experiments, potentially explaining the similar performance across methods.
However, training on similar problems may decrease generalization to out-of-distribution problems. 
This will be demonstrated in \Cref{subsec:res_large_meshes}, where accuracy decreases for an adapted geometry.

The baseline \ac{piggn} model based on \ac{gccn} performs significantly worse for the 2D experiments, compared to the proposed \ac{pimgn} and the \ac{ddmgn} baseline. 
For the 2D linear problem, the error is even more than an order of magnitude higher. 
In addition, the accuracy of the model is highly uncertain, which makes the training unreliable. 
The performance in the 3D mixed boundary task is slightly decreased, but still comparable. As for the \acp{ddmgn}, it will be demonstrated in \Cref{subsec:res_large_meshes} that the trained models generalize worse to large meshes.
This shows that the proposed \ac{mgn} architecture plays an important role in the generalization capabilities and reliability of \acp{pimgn}, allowing the approach to perform well on non-stationary problems of unseen arbitrary meshes, material distributions and process conditions. 

Overall, the proposed \ac{pimgn} alleviates the expensive data creation process required for \acp{ddmgn} while showing improved accuracy and consistency on different tasks. Further, the performance of the \ac{pimgn} increases up to an order of magnitude, compared to the physics-informed baseline \ac{piggn}.

\subsection{Generalization to large unseen meshes} \label{subsec:res_large_meshes}

\begin{figure}[!htb]
    \centering
    \includegraphics[]{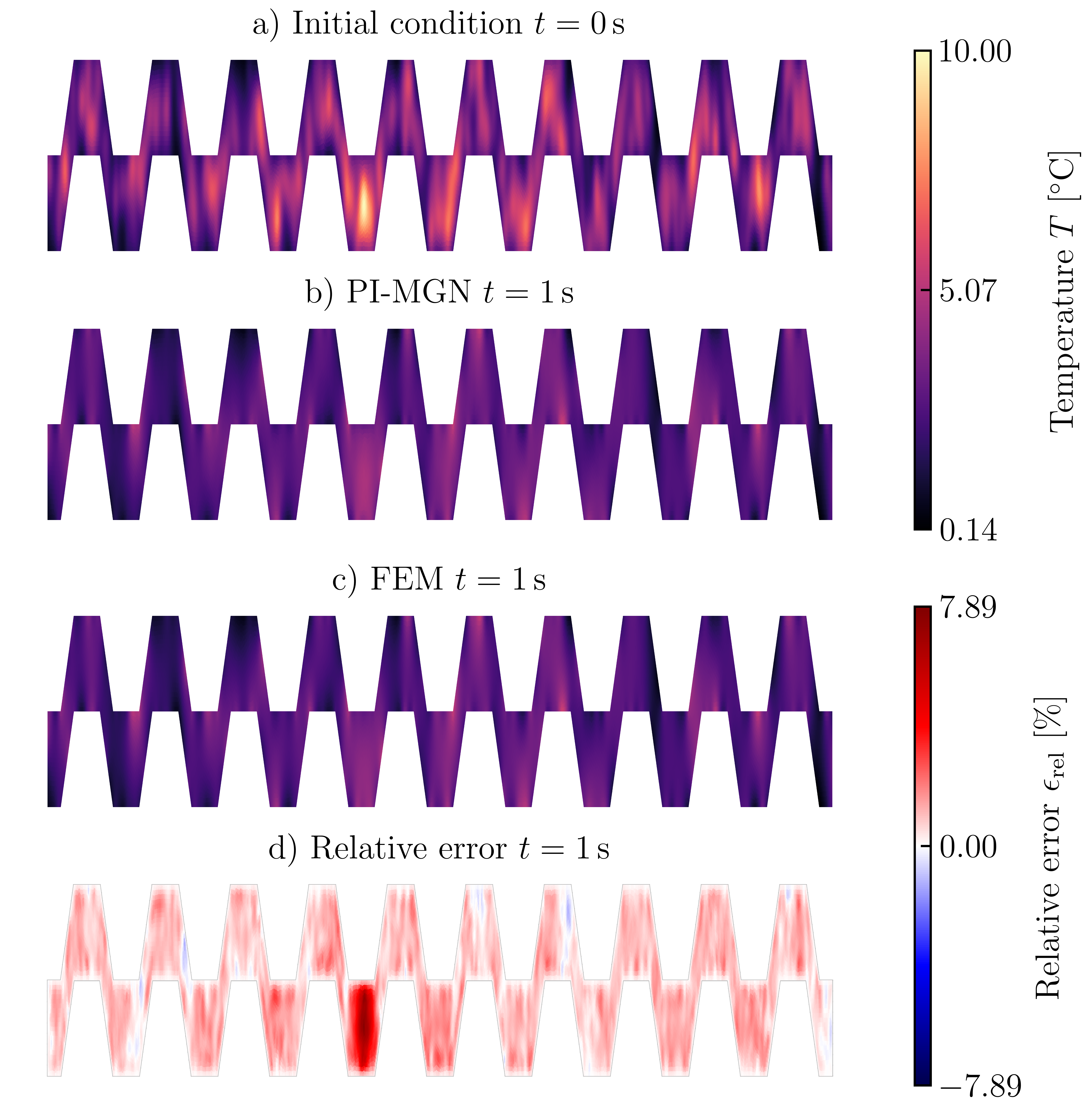}
    \caption{2D linear heat diffusion experiment of \Cref{subsec:exp_generalization} on a large mesh consisting of $10$ components. The part geometry and the random initial condition of \Cref{fig:UHDL_large} a) was not part of the training. \Cref{fig:UHDL_large} c) shows the solution of a \ac{pimgn}, which was trained on the small meshes of \Cref{subsec:exp_small_meshes}. The deviation from the ground truth \ac{fem} solution (cf. \Cref{fig:UHDL_large} c)) is depicted in \Cref{fig:UHDL_large} d). For visualization purposes, the image is not to scale}
    \label{fig:UHDL_large}
\end{figure}

\paragraph{Qualitative comparison to the \ac{fem}} The pre-trained models from \Cref{subsec:res_small_meshes} are applied to predict the time evolution on two large mesh experiments from \Cref{subsec:exp_generalization} without any retraining or modification. 
The results are compared qualitatively to the \ac{fem}. 
\Cref{fig:UHDL_large} a) shows the initial condition for a randomly selected model of the $5$ trained models. 
Even though the corrugated sheet mesh differs greatly from the L-shapes seen during training, the \ac{pimgn} accurately predicts (cf. \Cref{fig:UHDL_large} b)) the solution at the last time step when compared to the solution of the \ac{fem} in \Cref{fig:UHDL_large} c). \Cref{fig:UHDL_large} d) depicts the relative error between the \ac{pimgn} and the \ac{fem}, showing that the \ac{pimgn} tends to overestimates the solution. 
The highest error occurs at a point with high temperatures and temperature gradients, which are significantly larger than those observed during training. 

\begin{figure}[!htb]
    \centering
    \includegraphics[]{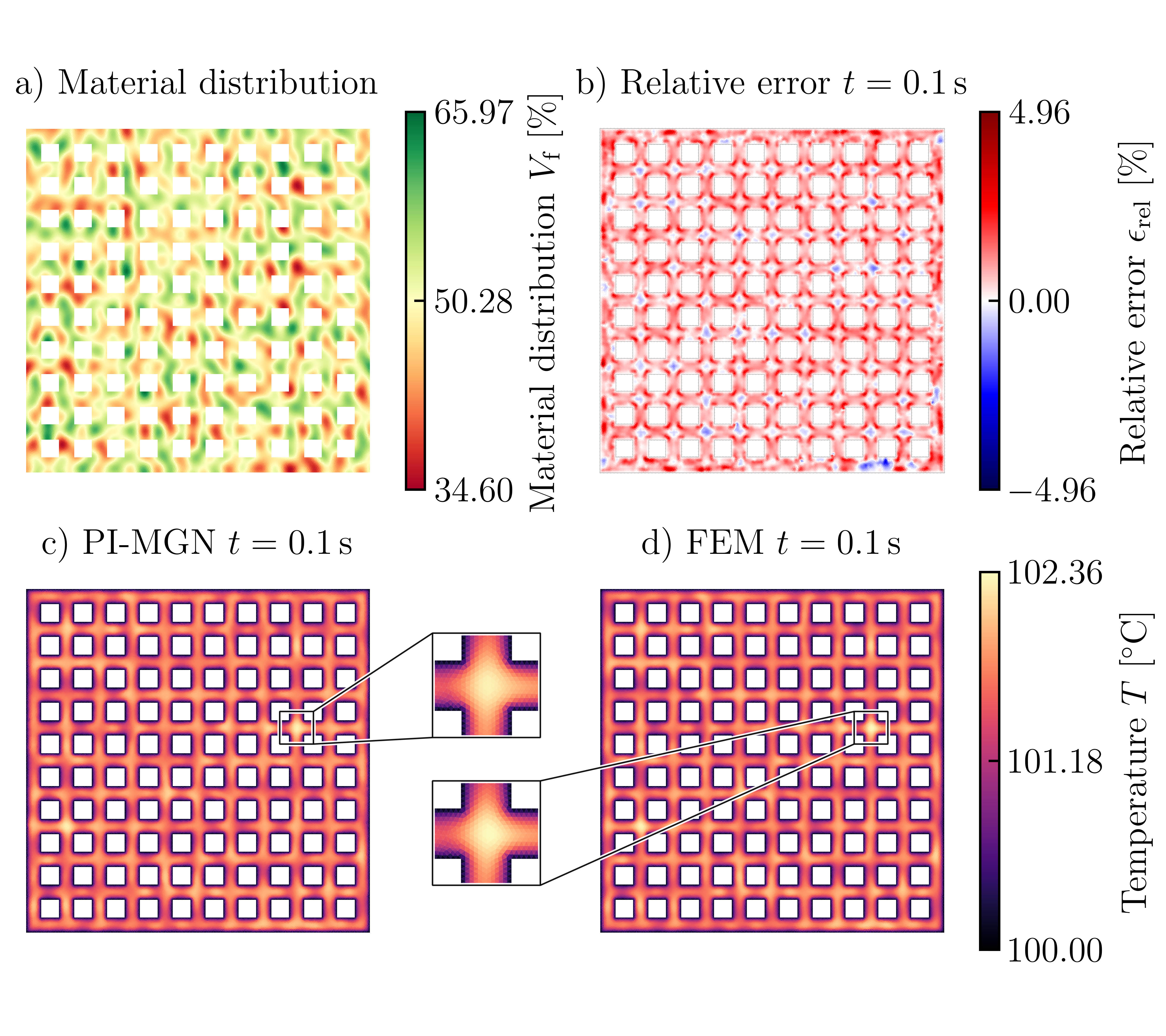}
    \caption{A trained \ac{pimgn} predicts the solution of the nonlinear experiment of \Cref{subsec:exp_generalization} on an large grid mesh. These \ac{pimgn} is only trained on small polygon meshes. The large part contains an inhomogeneous material distribution, which is depicted in \Cref{fig:UHDNL_large} a). \Cref{fig:UHDNL_large} c) shows the approximation of the \ac{pimgn}, \Cref{fig:UHDNL_large} d) the ground truth and b) the relative error the \ac{pimgn} solution.}
    \label{fig:UHDNL_large}
\end{figure}

The material distribution of the grid structure is depicted in \Cref{fig:UHDNL_large} a). 
The solution of the randomly chosen \ac{pimgn} in \Cref{fig:UHDNL_large} c) is in good agreement with the \ac{fem} solution of \Cref{fig:UHDNL_large} d). 
As before, the local temperature increases depend on the surrounding material distribution and are especially prominent at the crossings of the grid structure. 
The \ac{pimgn} provide accurate predictions and resolves fine local temperature changes. 
Interestingly, this is in contrast to classical \acp{pinn}, which have been reported to have a spectral bias toward learning low frequencies and may not be able to resolve such fine temperature changes \citep{caoUnderstandingSpectralBias2021, waheedKroneckerNeuralNetworks2022}. 
\Cref{fig:UHDNL_large} b) shows that the \ac{pimgn} tends to overestimate the temperature at the bars and to underestimate it at the crossings.

\paragraph{Comparison to baseline models} 

\begin{figure}[!htb]
    \centering
    \includegraphics[]{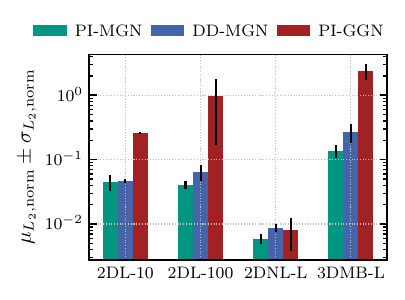}
    \caption{Performance of the \acp{pimgn} on the large mesh experiments of \Cref{subsec:exp_generalization} compared to \acp{ddmgn} and to \acp{piggn}. The mean $\mu_{L_2,\text{norm}}$ and the standard deviation $\sigma_{L_2,\text{norm}}$ of the normalized $L_2$ error $\epsilon_{L_2,\text{norm}}$ is calculated for five training repetitions per model and experiment. For each large mesh experiment, \Cref{fig:comp_base_lm} displays the mean $\mu_{L_2,\text{norm}}$ (bar) and the standard deviation $\sigma_{L_2,\text{norm}}$ (black line on bar) of the three models. \\}
    \label{fig:comp_base_lm}
\end{figure}

\Cref{fig:comp_base_lm} compares the \ac{pimgn}, \ac{ddmgn} and \ac{piggn} accuracy on the four large mesh experiments: The 2D linear heat diffusion with a corrugated sheet part, which consists of $10$ repeating components ('\textit{2DL-10}') respectively $100$ repeating components ('\textit{2DL-100}'), the 2D nonlinear and inhomogeneous heating of the grid structure ('\textit{2DNL-L}') and the 3D heating with mixed boundary conditions of the long hollow cylinder ('\textit{3DMB-L}'). As before, the models from \Cref{subsec:res_small_meshes} are re-used without any further training or modification. \Cref{fig:comp_base_lm} shows the mean $\mu_{L_2,\text{norm}}$ and standard deviation $\sigma_{L_2,\text{norm}}$ of the normalized $L_2$ error across the $5$ repetitions.

 The \acp{ddmgn} show comparable performance for the 2D linear heat diffusion task of the corrugated sheet mesh with 10 components.
 The mean error $\mu_{L_2,\text{norm}}$ is slightly higher and the standard deviation $\sigma_{L_2,\text{norm}}$ lower. For the other three problems, the \acp{pimgn} outperform the \acp{ddmgn}.
 Hence, the physics-informed training of \acp{mgn} not only saves on expensive data creation but also improves generalization to larger meshes.
 A reason for that could be that the physics-informed training has to explore the solution during training, because no ground truth is known.
 The solution approaches the ground truth during training, but always fluctuates due to the model inaccuracy.
 Therefore, the model must predict a slightly different solution with respect to the imprecise output of the previous time step, which can help to stabilize the model rollout. 

\paragraph{Ablation studies} The ablation studies conducted in \ref{chap:appendix_ablation_studies} demonstrate the effectiveness of each component of the \ac{pimgn} architecture in the small mesh experiments and its ability to generalize to larger, unseen meshes. For the training on small meshes, the \ac{pimgn} shows the highest reliability in predicting accurate results across all experiments. Time bundling and global features are important components to achieve this reliability. Replacing the relative positional encoding with absolute positions as well as noise free training prove to be competitive alternatives. However, the results on the large meshes suggest that \ac{pimgn} require relative encoding and robustness to noise to efficiently generalize beyond their training distribution.

\begin{figure}[!htb]
    \centering
    \includegraphics[]{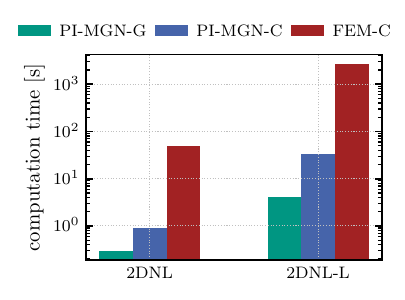}
    \caption{Speed comparison of the \acp{pimgn} to the \ac{fem} on the 2D nonlinear and inhomogeneous heating for the small meshes ('\textit{2DNL}') and large meshes  ('\textit{2DNL-L}'). \Cref{fig:comp_speed} compares the computation time of the \ac{pimgn} on a GPU ('\textit{\ac{pimgn}-G}') to its computation time on a CPU ('\textit{\ac{pimgn}-C}') and to the \ac{fem} on a CPU ('\textit{\ac{fem}-C}'). The GPU results are calculated on an NVIDIA\textsuperscript{\textregistered} GeForce\textsuperscript{\textregistered} RTX 2080 and the CPU results on an Intel\textsuperscript{\textregistered} Xeon\textsuperscript{\textregistered} CPU E5-2680 v4 @ 2.40GHz. \\}
    \label{fig:comp_speed}
\end{figure}

\paragraph{Speed comparison to the \ac{fem}} Finally, \Cref{fig:comp_speed} compares the computation time of the \acp{pimgn} to the \ac{fem} for the 2D nonlinear and inhomogeneous heating. The results of the \acp{pimgn} are calculated on a GPU (NVIDIA\textsuperscript{\textregistered} GeForce\textsuperscript{\textregistered} RTX 2080) and on a CPU (Intel\textsuperscript{\textregistered} Xeon\textsuperscript{\textregistered} CPU E5-2680 v4 @ 2.40GHz), while the \ac{fem} results are only calculated on the CPU.

The \acp{pimgn} require on average only about \SI{0.3}{s} on the GPU and \SI{0.9}{s} on the CPU for the small meshes compared to approximately \SI{50}{s} with the \ac{fem} on the same CPU.
For the large mesh grid, the \acp{pimgn} output the solution in about \SI{4}{s} on the GPU and in about \SI{33.8}{s} on the CPU while the \ac{fem} solver requires \SI{45}{min} on the CPU. \\

To sum up, the \acp{pimgn} trained on small meshes enable the most accurate and reliable predictions on large meshes over all considered learned models, while achieving speedups of several orders of magnitude compared to the \ac{fem}.

\section{Conclusion and Outlook}
\label{chap:conclusion_and_outlook}

This work introduces \acfp{pimgn} for fast, accurate and reliable \acf{fem} simulations of non-stationary problems on arbitrary mesh-geometries, material distributions and process conditions. \acp{pimgn} efficiently combine \acfp{pinn} and \acfp{mgn} to overcome the time-intensive and cumbersome data creation process required for \acfp{ddmgn} while improving accuracy and generalization compared to existing \acp{pinn}. During training, the \ac{fem} equations are discretized and evaluated per mesh element, allowing to compute the physics-informed loss for multiple time steps and nonlinear problems directly and in parallel. The state of the art \acp{mgn} \citep{pfaffLearningMeshBasedSimulation2020} are extended with additional methods such as time-bundling and global graph features, all of which empirically improve performance.

The \acp{pimgn} predict accurate solutions in good agreement with numerical \ac{fem} solvers for unseen arbitrary parts with inhomogeneous materials under changed process conditions. For a 2D heat diffusion problem, the \acp{pimgn} precisely predict the thermal cool down of unseen parts under random initialized temperature distributions. Further, the \acp{pimgn} learn to account for random material distributions in a part when predicting a nonlinear 2D heat release process. Finally, the approach accurately predicts the heating for 3D parts with mixed boundary conditions as a result of heat flow at the boundary. 

The introduced physics-informed training of the \acp{mgn} is compared to a data-driven training approach and to a physics-informed neural \ac{fem} solver. Even though \acp{pimgn} require no simulation data during training, they perform on par with or better than the data-driven variant in terms of accuracy on all experiments. This advantage of the physics-based training is even more significant when generalizing to larger parts during evaluation. In contrast to the data-driven training, the \acp{pimgn} reliably predict accurate simulations on large unseen meshes when trained only on small and generic meshes. The existing physics-informed neural \ac{fem} solvers in literature use architectures and parameterizations that are shown to perform worse in all experiments for both small and large meshes, highlighting the importance of the proposed \acp{pimgn}. An ablation study reveals the individual importance of several components of the \acp{mgn}, such as time bundling and global features.  

Overall, the \acp{pimgn} prove to be a reliable and accurate solver for the non-stationary and nonlinear \ac{fem} with strong generalization to unseen processes with new meshes consisting of inhomogeneous materials. In addition, the \acp{pimgn} are significantly faster and fully differentiable, making them a promising option for inverse design optimization. Here, the \acp{mgn} could also be extended to a Bayesian formulation to respect the model uncertainty during optimization. Finally, \acp{pimgn} could be applied to speed-up simulations with other challenging nonlinearities, such as contact or large deformation simulations.

\section{Acknowledgement}
\label{chap:acknowledgment}
This work has been performed in subprojects T2 and M2 of the DFG AI Research Unit 5339, funded by the Deutsche Forschungsgemeinschaft (DFG, German Research Foundation) – 459291153. This work is also part of the Heisenberg project ”Digitalization of fiber-reinforced polymer processes for resource-efficient manufacturing of lightweight components”, funded by the DFG (project no. 455807141). The authors thank the German Research Foundation for its financial support.

\section{Declaration of Competing Interest}
\label{chap:competiting_interest}
The authors declare that they have no known competing financial interests or personal relationships that could have appeared to influence the work reported in this paper.






\bibliographystyle{bibliography}
\bibliography{2024_CMAME_PI_MGN}

\newpage
\appendix
\section{Experiments}
\label{chap:appendix_A}
\newcolumntype{b}{X}
\newcolumntype{l}{>{\centering\arraybackslash\hsize=.9\hsize}X}
\newcolumntype{s}{>{\centering\arraybackslash\hsize=.6\hsize}X}

\newcolumntype{b}{X}
\newcolumntype{s}{>{\centering\arraybackslash\hsize=.3\hsize}X}
\newcolumntype{v}{>{\centering\arraybackslash\hsize=.5\hsize}X}
\newcolumntype{u}{>{\centering\arraybackslash\hsize=.3\hsize}X}

\subsection{Experiment 1: 2D linear heat diffusion} 
\label{subsec:apx_UHDL}

The first experiment (cf. \Cref{fig:experiments} a)) considers a heat diffusion problem on a 2D L-Shaped domain, setting the source term $q(T)=0$ in \Cref{eqn:pde_parabolic} and the thermal diffusivity $\alpha$ to a constant value. The initial condition is described by a sum of $N_{\text{ic}}$ Gaussians
$$
T^0(\mathbf{x})= \sum_{i=0}^{N_{\text{ic}}} a_i \exp \left(-\left(\frac{(x-x_{0,i})^2}{2\sigma_{x,i}}+\frac{(y-y_{0,i})^2}{2\sigma_{y,i}}\right) \right) \text{.}
$$
The factors $a_i$ and the standard deviations $\sigma_{x,i}, \sigma_{y,i}$ are randomly selected from uniform distributions. The means $x_{0,i},y_{0,i} \in \Omega \cup \partial \Omega$ are randomly selected node positions. The initial temperature on the boundary is fixed during the whole simulation, i.e., \mbox{$
\overline{T} = T^0(\mathbf{x})\text{,}~ \forall \mathbf{x} \in \Gamma
$}. Therefore, the problem contains no Neumann boundary condition, i.e. $\partial \Omega_{\mathcal{N}} = \{\}$.

The 2D L-shaped domain is defined as the difference of two rectangles. The large one has length $L$ and height $H$ and the small rectangle the length \mbox{$L_{\text{s}} = a_{\text{L}} L$} and the height $H_{\text{s}} = a_{\text{H}} H$, where $L$, $H$, $a_{\text{L}}$ and $a_{\text{H}}$ are selected randomly, cf. \Cref{fig:experiments} a). The small rectangle is randomly placed at one of the corners of the large domain and subtracted from the large rectangle to yield the final L-shaped domain. A total amount of $100$ L-shaped domains are created and remain the same for all five seeds. However, mesh selection and initial conditions are still randomized over seeds for training and validation. The exact definitions of the constants and the distributions of the randomly selected values are listed in \cref{table:experiments_values_uhdl}.

\begin{table}[!htb]
\caption{Constants and randomly defined values of Experiment 1, the 2D linear heat diffusion experiment.} 
{
    \centering
    \small
    \begin{tabularx}{\textwidth}{ b  s v u  }
            
            \hline
            \textbf{Property} & \textbf{Symbol} & \textbf{Value} & \textbf{Unit} \\
            \hline 
            
            \textit{Initial-boundary value problem} &  & & \\
            
             Source term & $q$ & 0 & \si{K/s}\\

            Thermal diffusivity & $\alpha$ & $\num{5e-2}$ & $\si{m^2/s}$\\

             Number of Gaussians & $N_{\text{ic}} $ & 10 & - \\

             Summand weight &  $a_i$ & $\sim \mathcal{U}(0.5,1)$ & \si{\degreeCelsius}  \\

             Standard deviation $x$-direction &  $\sigma_{x,i}$ & $\sim \mathcal{U}(1/12,1/6)$ & \si{m^2} \\

            Standard deviation $y$-direction&  $\sigma_{y,i}$ & $\sim \mathcal{U}(1/12,1/6)$ &  \si{m^2} \\

            \textit{Geometry - L-shape} &  & & \\

            Length &  $L$ & $\sim \mathcal{U}(0.5,1)$ & \si{m} \\

            Height &  $H$ & $\sim \mathcal{U}(0.5,1)$ & \si{m}  \\

            Length factor &  $a_{\text{L}}$ & $\sim \mathcal{U}(1/3,2/3)$ & - \\

            Height factor &  $a_{\text{H}}$ & $\sim \mathcal{U}(1/3,2/3)$ & - \\

            \hline
            
    \end{tabularx}\par

}
\label{table:experiments_values_uhdl}
\end{table}

\subsection{Experiment 2: 2D nonlinear and inhomogeneous heating} \label{subsec:apx_UHDNL}

This experiment (cf. \Cref{fig:experiments} b)) contains a nonlinear source term 
\begin{equation} \label{eqn:heat_source}
q(T) = q(T, t, \mathbf{x}) = \tilde{q}_0(\mathbf{x}) \exp{\left(- C T t / t_0 \right)} C  T
\end{equation}
in \Cref{eqn:pde_parabolic},
where $C = 8/T_0$, $\tilde{q}_0(\mathbf{x}) = q_{0}( 1- V_{\text{f}}(\mathbf{x}))$ and $t_0$, $T_0$ and $q_0$ are constant. The factor $\tilde{q}_0$ depends on an inhomogeneous material property $V_{\text{f}}$
\begin{equation*}
V_{\text{f}}(\mathbf{x}) = V_{\text{f,0}} + \sum_{i=1}^{N_f} a_i \sin \left( k_{x,i}  x + d_{x,i} \right) \sin \left( k_{y,i} y+ d_{y,i} \right) \text{.}
\end{equation*}

These material distributions are remotely inspired by inhomogeneities  which may occur in press forming processes of short fibre reinforced composites. The material field contains $N_f$ functions, each depending on a product of sine waves in $x$ and $y$-direction. The factor $a_i$, the wavenumbers $k_{x,i}$ and $k_{y,i}$ as well as function shifts  $d_{x,i}, d_{y,i}$ are randomly chosen from uniform distributions, while $V_{\text{f,0}}$ is a constant.
Additionally, the diffusion coefficient 
\begin{equation} \label{eqn:uhdnl_thermal_diffusivity}
\alpha = \alpha(\textbf{x}) = 1/\left(V_{\text{f}}(\mathbf{x})/\alpha_{\text{f}} + (1-V_{\text{f}}(\mathbf{x}))/\alpha_{\text{m}}\right)
\end{equation}
also depends on the inhomogeneous material distribution $V_{\text{f}}(\mathbf{x})$ as well as on the constants \mbox{$\alpha_{\text{f}}$} and $\alpha_{\text{m}}$. The temperature at the initial condition $T^0$ and at the whole boundary $\overline{T}$ is set to \mbox{$T^0 = \overline{T} = T_0$}.
The considered domains are convex polygons. These are constructed by uniformly sampling $7$ points $x_{\text{P}}$ and $y_{\text{P}}$ in the 2D domain $\Omega = (0,L) \times (0,H)$, cf. \Cref{fig:experiments} b). Subsequently, the convex hull of the set of the seven points is chosen as the domain, yielding a convex polygon with $3$ to $7$ boundary points. \Cref{table:experiments_values_uhdnl} lists the values of the constants and distributions of the random quantities.

The polygons and material distributions $V_{\text{f}}(\mathbf{x})$ of the $100$ problems are randomly selected in a pre-processing step and fixed for all training seeds. The problem selection of the $75$ training and $25$ validation is randomized across the five seeds.
As the heat source $ q(T, t, \mathbf{x})$ depends on time and temperature, the temperature T is added as a node feature and the time is added as both, a global and a node feature. In addition, the material property, evaluated at the nodes, is added as a node feature. 

To obtain a ground truth solution for this nonlinear equation from a numerical solver, the equation is linearized and solved in an inner loop for each time step for a total number of $100$ inner iteration or until the solution is converged, depending on what happens first.

\begin{table}[!htb]
 \caption{Constants and randomly defined values of Experiment 2, the 2D nonlinear and inhomogeneous experiment.} 
{
    \centering
    \small            

    \begin{tabularx}{\textwidth}{ b  s v u  }

            \hline
            \textbf{Property} & \textbf{Symbol} & \textbf{Value} & \textbf{Unit} \\
            \hline

            \textit{Initial-boundary value problem} &  & & \\

            Reference time & $t_0$ & $1$ & \si{s} \\ 

            Reference temperature & $T_0$ & $100$ & \si{\degreeCelsius} \\ 

            Scaling factor & $q_0$ & $20$ & $\si{K/s}$ \\ 
                       
            Number of functions & $N_f$ & $10$ & - \\
            
            Summand weight & $a_i$ & $\sim \mathcal{U}(0,1/20)$ & - \\

            Wavenumber $x$-direction & $k_{x,i}$ & $  \sim \mathcal{U}(0,8 \pi)$ & \si{1/m} \\
            Wavenumber $y$-direction & $k_{y,i}$ & $ \sim \mathcal{U}(0,8 \pi)$ & \si{1/m} \\

            Wave shift $x$-direction & $d_{x,i}$ &   $\sim \mathcal{U}(0,2 \pi)$ & - \\
            Wave shift $y$-direction & $d_{y,i}$ & $\sim \mathcal{U}(0,2 \pi)$ & - \\

            Median material property & $V_{\text{f,0}}$ & $0.5$ & - \\

            Thermal diffusivity fibre  & $\alpha_{\text{f}}$ & $0.1$ & $\si{m^2/s}$ \\

            Thermal diffusivity matrix  & $\alpha_{\text{m}}$ & $0.01$ & $\si{m^2/s}$ \\

            \textit{Geometry - Convex polygon} &  & & \\

            Geometry length &  $L$ & $1$ & \si{m} \\

            Geometry height &  $H$ & $1$ & \si{m}  \\

            \hline
            
\end{tabularx}\par

}
\label{table:experiments_values_uhdnl}
\end{table}       

\subsection{Experiment 3: 3D heating with mixed boundary conditions} \label{subsec:apx_UHDL3D}
In the following, the \ac{pde} of \ref{subsec:apx_UHDL} is considered without a heat source $q(T) = 0$ for a 3D hollow cylinder mesh with mixed boundary conditions (cf. \Cref{fig:experiments} c)).

The hollow cylinder has the length $L$, the outer radius $R_{\text{o}}$ and the inner radius \mbox{$R_{\text{i}} = a_{\text{i}} R_{\text{o}}$}, where  $L$, $R_{\text{o}}$ and the fracture $a_{\text{i}}$ are randomly selected values.
At the beginning, the temperature is $T^0=T_0$ across the domain. The temperature at the left and right cylinder boundary $\Gamma_{\text{l,r}} =  \{\mathbf{x} = (x,y)\in \partial \Omega: x = 0, L\} $  is also fixed to $\bar{T} = T_0$. A constant Neumann boundary condition is defined at the boundary surface at the inner side of the cylinder, i.e. at $\partial \Omega_{\text{i}} =  \{\mathbf{x} = (x,y) \in \partial \Omega: y = R_{\text{i}}\}$ as $ \alpha h_{\mathcal{N}} = h_0$. At the outer side of the cylinder $\partial \Omega_{\text{o}} =  \{\mathbf{x} = (x,y) \in \partial \Omega: y = R_{\text{o}}\}  $ an adiabatic Neumann boundary condition $ \alpha h_{\mathcal{N}} = 0$ is considered. The thermal diffusivity $\alpha$ is constant.

The initial learning rate is changed to \num{1e-4}, but still decays to \num{1e-5}. The $100$ meshes of the 3D hollow cylinders are the same for all five training seeds, but the part selection for the training and validation remains random for the individual seeds.  
The inner heat $h_0$ of the Neumann boundary condition is added as an additional node feature. The exact constants and the randomly selected value definitions are given in \cref{table:experiments_values_uhdl3D}.

\begin{table}[!htb]
\caption{Constants and randomly defined values of Experiment 3, the 3D heating experiment with mixed boundary conditions.} 
{
    \centering
    \small
    \begin{tabularx}{\textwidth}{b  s v u  }

            \hline
            \textbf{Property} & \textbf{Symbol} & \textbf{Value} & \textbf{Unit} \\
            \hline 
            
            \textit{Initial-boundary value problem} &  & & \\ 
            
            Source term & $q$ & 0 & \si{K/s}\\
            Reference temperature & $T_0$ & 0 &  \si{\degreeCelsius} \\ 
            Thermal diffusivity & $\alpha$ & $\num{1}$ & $\si{m^2/s}$\\

            Inner heat constant & $h_0$ & 1 & $\si{m ~K/s}$ \\

            \textit{Geometry - Hollow cylinder} &  & & \\

            Length &  $L$ & $\sim \mathcal{U}(4,5)$ & \si{m} \\

            Outer radius &  $R_{\text{o}}$ & $\sim \mathcal{U}(0.8,1)$ & \si{m} \\
            
            Fracture radius & $a_{\text{i}}$ & $\sim \mathcal{U}(0.6,0.8)$ & \si{m} \\

            \hline

\end{tabularx}\par

}
\label{table:experiments_values_uhdl3D}
\end{table}         

\Cref{table:graph_feauters} provides an overview of the node, edge and global features used in the experiments as well as the output values of the \acp{pimgn}. 

\begin{table}[!htp]
\caption{Input and output features used in each experiment, including node, edge, and global input features, as well as the output features on the nodes.} 
{
    \centering
    \small
    \begin{tabularx}{\textwidth}{ b l l  s s}
    
            \hline
            \textbf{Experiment}  & \textbf{Node} & \textbf{Edge} & \textbf{Global} & \textbf{Outputs} \\
            \hline

             2D Linear  & $n_\text{t}$ & $\mathbf{x}_{vu}$, $|\mathbf{x}_{vu}|$, $T_{vu}$ & - & $T$\\

            2D Nonlinear   & $n_\text{t}$, $t$, $T$, $V_\text{f}$ & $\mathbf{x}_{vu}$, $|\mathbf{x}_{vu}|$, $T_{vu}$ & $t$ & $T$\\

            3D mixed BCs  & $n_\text{t}$, $h_0$   & $\mathbf{x}_{vu}$, $|\mathbf{x}_{vu}|$, $T_{vu}$ & - & $T$\\
            
            \hline
            
    \end{tabularx}\par
}
\label{table:graph_feauters}
\end{table}

\subsection{Large mesh problems of the experiments}
\label{subsec:apx_generalization}

For the first two problems, the models have to predict the 2D linear heat diffusion of \ref{subsec:apx_UHDL} for large corrugated sheets, which consist of repeating components of length $L$ and thickness $D$. The first problem considers a mesh consisting of $N_\text{c1} = 10$ repeating components with $\num{10840}$ total mesh elements and the second of $N_\text{c2} = 100$ components with $\num{107592}$ elements. The number of Gaussians for the initial condition is increased to $N_{\text{ic}} = 1000$. As a result, the density of the Gaussians in the initial state in the two parts varies by a factor of $10$. These evaluation problems are multiple magnitudes larger larger than anything seen during training, as the training process for all methods uses meshes with $155$ to $524$ elements and $N_{\text{ic}} = 10$ Gaussians.

The third problem considers the 2D nonlinear heating of \ref{subsec:apx_UHDNL} for a grid structure. The grid part is a large rectangle with the length and height $L$ and contains $N_\text{h} = 100$ rectangle holes equally distributed along the length and height direction and with a distance of $d_\text{h}$. The mesh of the grid contains $\num{23296}$ elements, while the training meshes contain between $148$ and $ 1122$ elements. 

The last problem considers the 3D heatup with mixed boundary conditions of \ref{subsec:apx_UHDL3D} for a long 3D hollow cylinder, whose length $L$ is 40-50 times larger than that of the training meshes, while the outer radius $R_\text{o}$ and the inner radius multiplier $a_{\text{i}}$ correspond to the mean value of the training meshes. The other settings remain the same as in \ref{subsec:apx_UHDL3D}. 

The added constants and distributions of the randomly selected values for these experiments are listed in \cref{table:experiments_values_large_meshes}.

\begin{table}[H]
\caption{Constants and randomly defined values of the large mesh experiments.} 
{
    \centering
    \small
    \begin{tabularx}{\textwidth}{ b  s v u }

            \hline
            \textbf{Property} & \textbf{Symbol} & \textbf{Value} & \textbf{Unit} \\
            \hline 
            
            \textbf{Experiment 1: 2D Linear} &  & & \\
            
            \textit{Initial-boundary value problem} &  & & \\
            
             Number of Gaussians & $N_{\text{ic}} $ & 1000 & - \\
            
            \textit{Geometry - Corrugated sheet} &  & & \\

             Component length & $L$ & $3$ & \si{m} \\
             Component thickness & $D$ & $0.5$ & \si{m} \\
            Number of components part 1 & $N_\text{c1}$ & 10 & - \\
             Number of components part 2 & $N_\text{c2}$ & 100 & - \\

            \hline

            \textbf{Experiment 2: 2D Nonlinear} &  & & \\

            \textit{Geometry - Large rectangle} &  & & \\
            Length / Height & $L$& $4$ & \si{m} \\
            Number of holes & $N_\text{h}$& $100$ & - \\
             Distance holes & $d_\text{h}$& $0.175$ & \si{m} \\

             \hline
             \textbf{Experiment 3: 3D mixed BCs} &  & & \\
             
            \textit{Geometry - Long hollow cylinder} &  & & \\
            Length & $L$ & $200$ & \si{m} \\
            Outer radius &  $R_{\text{o}}$ & $0.9$ & \si{m} \\
            
            Fracture radius & $a_{\text{i}}$ & $0.7$ & \si{m} \\

            \hline
            
    \end{tabularx}\par

}
\label{table:experiments_values_large_meshes}
\end{table}


\section{Ablation studies}
\label{chap:appendix_ablation_studies}
\newcolumntype{k}{>{\centering\arraybackslash\hsize=.9\hsize}X}

The contribution and effectiveness of individual components of the proposed \ac{pimgn} architecture is investigated in an ablation study. In each ablation, a component of the architecture is removed and, if necessary, replaced by a suitable alternative. 

\paragraph{Training on small meshes} The first ablation uses absolute positions as node features instead of a relative positional encoding at the edges of the input graph of the \ac{pimgn} ('\textit{abs. pos.}'). Next, the training noise is left out, instead using unmodified predictions as input for the next \ac{mgn} step ('\textit{w/o noise}'). Then, global features are removed in \Cref{eqn:mpn} of the \ac{pimgn} ('\textit{w/o global}') and the \ac{cnn} decoder of the model is replaced by an \ac{mlp} decoder ('\textit{MLP decoder}'). The last two ablations use different time bundling sizes $N_{\text{TB}}$, in particular $N_{\text{TB}} = 10$ and $N_{\text{TB}} = 50$ ( '\textit{TBS-10}' respectively '\textit{TBS-50}') instead of $N_{\text{TB}} = 20$. These ablations investigate how accuracy changes when the number of model calls is doubled or approximately halved in the experiments.

\Cref{table:ablation} shows the results for \acp{pimgn} and all ablations. The table contains the mean $\mu_{L_2,\text{norm}}$ and standard deviation $\sigma_{L_2,\text{norm}}$ of the error $\epsilon_{L_2,\text{norm}}$ over five training repetitions for 25 unseen problems per experiment. 

\begin{table}[!htb]

\caption{Ablation study of the \acp{pimgn} for five training repetitions per method and experiment, considering 25 unseen evaluation problems. The table displays $\mu_{L_2,\text{norm}} \pm \sigma_{L_2,\text{norm}}$, where $\mu_{L_2,\text{norm}}$ is the mean and $\sigma_{L_2,\text{norm}}$ the standard deviation of the normalized $L_2$ error $\epsilon_{L_2,\text{norm}}$, each multiplied by \num{1e+3}. The top-performing methods are highlighted in bold: blue for the best, green for the second-best, and yellow for the third-best. \\}
{
\centering
\footnotesize
\begin{tabularx}{\textwidth}{ X k k k }

        \hline
        \textbf{Methods} & \textbf{2D Linear} & \textbf{2D Nonlinear} & \textbf{3D mixed BCs}\\
        \hline
        \ac{pimgn}  & \textbf{\textcolor{kitorange}{17.19 $\pm$ 1.13}} & \textbf{\textcolor{kitgruen}{0.73  $\pm$ 0.15}} & \textbf{\textcolor{kitorange}{86.07  $\pm$ 2.97}}\\
        
        abs. pos. & 17.73  $\pm$ 0.96 & 0.97  $\pm$ 0.13 & \textbf{\textcolor{kitblau}{47.57  $\pm$ 18.37}}\\

        w/o noise & \textbf{\textcolor{kitblau}{16.08  $\pm$ 15.53}} & 0.86 $\pm$ 0.16 & 88.53  $\pm$ 6.57 \\
        
        w/o global feature & \textbf{\textcolor{kitgruen}{16.37  $\pm$ 0.72}} & \textbf{\textcolor{kitblau}{0.68  $\pm$ 0.05}} & 162.71  $\pm$ 8.65\\
        
        MLP decoder & 173.52  $\pm$ 143.60 & 0.86  $\pm$ 0.15 & \textbf{\textcolor{kitgruen}{85.13 $\pm$ 4.38}}\\

        TBS-10 & 75.89  $\pm$ 125.50 & \textbf{\textcolor{kitorange}{0.77  $\pm$ 0.08}} & 90.94 $\pm$ 9.73\\
        
        TBS-50 & 21.80  $\pm$ 3.93 & 1.74  $\pm$ 0.06 & 88.37  $\pm$ 7.10\\

        \hline

\end{tabularx}\par

}
\label{table:ablation}
\end{table}

Only the proposed \acp{pimgn} architecture perform well in all experiments and consistently score within the top three models in terms of mean error $\mu_{L_2,\text{norm}}$. The absolute position encoding shows comparable performance in two of the three experiments and even outperforms the \acp{pimgn} in the 3D experiment of \ref{subsec:apx_UHDL3D}. However, the relative positional encoding is a more solid representation for generalization to geometries, which will be shown in \Cref{subsec:res_large_meshes}.  Removing the training noise leads to increased standard deviations in two of the three experiments, while the mean error remains comparable. Without global features, the prediction error slightly decreases for the 2D experiments, but almost doubles in the 3D case. The same applies for the \ac{mlp} decoder ablation \textit{MLP decoder} and the two time bundling size ablations \textit{TBS-10} and \textit{TBS-50}. All of them perform similar to the \ac{pimgn} in two of the experiments, but significantly worse in the third. These components of the architecture might not be necessary in all cases, but prove to be crucial for the reliability of the model. It is important to note that accuracy tends to decrease with fewer model calls (\textit{TBS-50}), even though less error accumulation should increase accuracy. This phenomenon may be attributed to poorer generalization.

\paragraph{Generalization to large unseen meshes}
Since training without noise and absolute position encoding showed the most competitive performance, these two ablations are compared to the proposed \acp{pimgn} architecture for generalization to large unseen meshes.
\Cref{table:large_meshes_abl} contains the results of the proposed \ac{pimgn} architecture, the \ac{pimgn} with absolute positional encoding ('\textit{abs. pos.}') and the noise-free variant ('\textit{w/o noise}'). Therefore, the pre-trained models from the small mesh ablations are applied on the four large mesh experiments. The values of \Cref{table:large_meshes_abl} are the mean $\mu_{L_2,\text{norm}}$ and standard deviation $\sigma_{L_2,\text{norm}}$ of the normalized $L_2$ error across the $5$ repetitions. 

\begin{table}[!htb]

\caption{Comparison of the \acp{pimgn} to the best performing variants of the small mesh ablations for the experiments on large meshes. The mean $\mu_{L_2,\text{norm}}$ and the standard deviation $\sigma_{L_2,\text{norm}}$ of the normalized $L_2$ error $\epsilon_{L_2,\text{norm}}$ is calculated for five training repetitions per model and experiment. The tables displays the values in the format $\mu_{L_2,\text{norm}} \pm \sigma_{L_2,\text{norm}}$ multiplied by $\num{1e+3}$. The best methods are highlighted in bold. \\}
{
\centering
\footnotesize
\begin{tabularx}{\textwidth}{ X c c c c }

        \hline
        \textbf{Methods} & \textbf{2DL-10} & \textbf{2DL-100} & \textbf{2DNL-L} &  \textbf{3DMB-L}\\
        \hline
        \ac{pimgn} & \textbf{\textcolor{kitblau}{45.42 $\pm$ 12.97}} & 41.06 $\pm$ 6.32 & \textbf{\textcolor{kitblau}{5.93 $\pm$ 1.12}} & \textbf{\textcolor{kitblau}{138.13 $\pm$ 33.55}} \\
                
        abs. pos. &  183.83 $\pm$ 82.99 & 285.54 $\pm$  78.13  & 6.33 $\pm$ 1.43 &  766.06 $\pm$ 116.25\\
        
        w/o noise & 48.53 $\pm$ 26.34 & \textbf{\textcolor{kitblau}{34.00 $\pm$ 1.82 }} &  6.85 $\pm$ 0.81 &  240.78 $\pm$ 141.71 \\
        \hline

\end{tabularx}\par
}
\label{table:large_meshes_abl}
\end{table}

The \acp{pimgn} have the lowest error in 3 out of 4 problems. Only the noise free training slightly outperforms the \acp{pimgn} for the 2D linear heat diffusion problem of the corrugated sheet with 100 components, but clearly struggles to solve the long 3D hollow cylinder problem. The absolute positional encoding only performs well on the nonlinear task with the large grid structure and fails to provide accurate predictions in the other three problems. Remarkably, the noise-free training and absolute positional encoding ablations show decreased performance for the seemingly small change of the 3D problem setup, which considers only a longer hollow cylinder.

\end{document}